\documentclass[journal,10pt,doublecolumn,singlespace]{IEEEtran}

\usepackage{cite,graphicx,subfig,multirow,array}

\usepackage{amsfonts,amssymb}
\usepackage[cmex10]{amsmath}
\usepackage{algorithmic}
\usepackage[ruled]{algorithm2e}	
\usepackage{epstopdf}
\usepackage{hyperref}
\usepackage{verbatim}
\usepackage[usenames,dvipsnames]{color}
\usepackage{soul}
\usepackage{tabularx}
\usepackage{makecell}
\usepackage{microtype}
\usepackage{booktabs}
\usepackage{textcomp}
\usepackage{animate}

\begin{document}
	
\pagenumbering{gobble}

\title{Temporal-MPI: Enabling Multi-Plane Images for Dynamic Scene Modelling via Temporal Basis Learning}

\author{\IEEEauthorblockN{Wenpeng Xing, Jie Chen, \textit{Member, IEEE}}

\thanks{W. Xing and J. Chen (correspondence author) are with the Department of Computer Science, Hong Kong Baptist University, Hong Kong. E-mails: \{cswpxing, chenjie\}@comp.hkbu.edu.hk.}

}

\markboth{Manuscript}
{\MakeLowercase{\textit{Xing et al.}}:}

\maketitle

\begin{abstract}
Novel view synthesis of static scenes has achieved remarkable advancements in producing photo-realistic results. However, key challenges remain for immersive rendering of dynamic scenes. One of the seminal image-based rendering method, the multi-plane image (MPI), produces high novel-view synthesis quality for static scenes. But modelling dynamic contents by MPI is not studied.
In this paper, we propose a novel Temporal-MPI representation which is able to encode the rich 3D and dynamic variation information throughout the entire video as compact temporal basis and coefficients jointly learned. Time-instance MPI for rendering can be generated efficiently using mini-seconds by linear combinations of temporal basis and coefficients from Temporal-MPI. Thus novel-views at arbitrary time-instance will be able to be rendered via Temporal-MPI in real-time with high visual quality. Our method is trained and evaluated on Nvidia Dynamic Scene Dataset. We show that our proposed Temporal-MPI is much faster and more compact compared with other state-of-the-art dynamic scene modelling methods.

\end{abstract}

\begin{figure*}[h]
    \centering
    \begin{tabular}{ccc}
            \animategraphics[autoplay,loop,width=4cm, ]{8}{figure/umbrella/frames/frame_0}{051}{54}
        &
            \animategraphics[autoplay,loop,width=4cm, ]{8}{figure/balloon_2-2/frames/frame_0}{000}{03}
         &
           \animategraphics[autoplay,loop,width=4cm, ]{8}{figure/dynamic_face/frames/frame_0}{000}{03}
    \end{tabular}
    \caption{Reconstruction quality demonstration of the proposed Temporal-MPI for the testing sequences from the Nvidia Dynamic Scene dataset \cite{yoon2020novel}. The dynamic visual effects can be viewed in Adobe PDF Reader.
    }
\end{figure*}

\section{Introduction}
Recent advancements on novel view synthesis have shown remarkable results on immersive rendering of static scenes using neural scene representations, such as Multi-plane Images (MPI) \cite{zhou2018stereo,Wizadwongsa2021NeX,mildenhall2019llff} and Neural Radiance Fields (NeRF) \cite{mildenhall2020nerf,chen2021mvsnerf}. Neural basis expansion \cite{Wizadwongsa2021NeX} and Plenoctree structures \cite{yu2021plenoctrees} have been recently proposed to further improve the rendering quality and efficiency. However, challenges still remain in modelling dynamic scenes, which require additional capacity to capture variations along time dimension.

To model dynamic contents, efforts have been made in training time-conditioned NeRF models \cite{li2021neuralsceneflow,li2022neural,pumarola2021d}.
Although photo-realistic view-synthesis results can be produced by these time-conditioned neural rendering methods, they normally require millions of ray-casting style queries
during rendering, resulting in serious rendering delay and low frame rate. So, there is a popular branch of research on improving the rendering efficiency of neural scene representations by extracting the learned content into compact data structure, such as tree-based structure \cite{yu2021plenoctrees}, or with occupancy priors \cite{liu2020neuralspv} stored for efficient sampling. Another line of image-based rendering research, the MPI, focuses on rendering real-world forward-facing contents. MPI is highly efficient for real-time rendering due to its pre-computed 2.5D RGB-$\alpha$ volumes.
In order to render dynamic scenes via MPI, pre-calculating and saving all time-instance MPIs is a straight-forward but engineering-oriented solution for time-space rendering. However, this method lacks temporal coherence and is expensive to save the bulky data incurred. \textit{3DMaskVol21} \cite{lin2021deep} renders an image at a given timestamp by fusing a background MPI and instantaneous MPI using a 3D mask volume, which takes temporal-coherent information learned to be the background MPI.
But generating these three volumes causes delay on rendering and heavy work-load on caching.
In comparison, our proposed method can generate arbitrary time-instance MPIs from one Temporal-MPI within mini seconds, which is much more efficient for real-time rendering and compact in storage.

In this paper, we propose a novel efficient representation for dynamic scenes, Temporal-MPI, for space-time immersive rendering. Different from previous methods \cite{li2021neuralsceneflow,lin2021deep,yoon2020novel} which rely on pre-trained optical flow model \cite{ilg2017flownet}, ground-truth background images \cite{lin2021deep}, pre-trained depth estimation model \cite{Ranftl2020Towards} or dynamic-static masks \cite{yoon2020novel} as additional premise, we aim at creating a self-contained pipeline.
In addition, our method does not need to explicitly store time-instance MPIs, which greatly decreases the requirement for storage space and being computationally efficient.

\section{Related Work}

\noindent\textbf{Novel view synthesis.}
Novel view synthesis is a long standing research issue that aims at synthesising novel views of a scene given arbitrary captured images, and has become one of the most popular classes of research topics in computer vision. Early researches on Light fields (LF) \cite{ng2005light} represented the scene as a 4D Plenoptic Function \cite{mcmillan1995plenoptic} $L(x, y, s, t)$, where $(x,y)$ represents spatial coordinates and $(s,t)$ represents angular coordinates.
The spatial-angular correlations embedded in LF images can be exploited for applications of depth estimation~\cite{2015Accurate,7847308}, super-resolution \cite{9204825} and novel view rendering \cite{srinivasan2017learning,LearningViewSynthesis}. With recent advancements in deep learning, different learning-based scene representations were proposed. Novel views can be synthesized from monocular input, they are \textit{SynSin20} \cite{wiles2020synsin}, \textit{3D Photo20} \cite{Shih3DP20}, \textit{WorldSheet21} \cite{hu2021worldsheet} and \textit{MPIs20}~\cite{single_view_mpi}. They share a common rationale, which is integrating the learning of geometry and appearance from rendering loss. The second branch of research is using multi-view inputs that allow machine learning models to reason the scene's geometry using epipolar geometry and triangulation, the scene can be learned as a volume representation either explicitly or implicitly. Neural/implicit volume representations can encode the scene as a continuous volumetric function, they are Deep Voxels \cite{sitzmann2019deepvoxels} and NeRF \cite{mildenhall2020nerf}. In addition to above continuous volumetric functions, a scene can be decomposed into a layered representation \cite{ldi}, they are MPI \cite{zhou2018stereo} and its
followers \cite{mildenhall2019llff,single_view_mpi,Wizadwongsa2021NeX,lin2021deep}.  Although these methods can produce photo-realistic results, they can only model and render static scenes. The next key step of view rendering is rendering dynamic scenes.

\noindent\textbf{Neural spatial and temporal embedding for novel-view synthesis.}
A successful novel view synthesis requires accurate modeling of a scene's geometry. Modeling the geometry of non-rigid scenes with dynamic contents are ill-posed, and were tackled by reconstructing dynamic 3D meshes where priors like temporal information \cite{agudo2015simultaneous,tomasi1992shape} or known
template configurations \cite{bartoli2015shape,moreno2012stochastic}. Yet, these methods require 2D-to-3D matches or 3D point tracks. Thus, limiting their applicability to real world scenes or simulated scenes with complex textures.

Under the context of space-time view synthesis, adding time parameters to the input of static scene's representations is a straightforward implementation. There are time-conditioned warping fields in \textit{D-NeRF21} \cite{pumarola2021d}, scene flow fields in \textit{NeuralFlow21} \cite{li2021neuralsceneflow} and radiance fields in \textit{Neural3DVideo21} \cite{li2022neural}. More specifically, \textit{D-NeRF21} \cite{pumarola2021d} added a time-conditioned deformation network to predict the time-dependent positional offsets to deform the canonical NeRF into a time-instance shape. \textit{NeuralFlow21}~\cite{li2021neuralsceneflow} used temporal photometric consistency to encourage the time-conditioned NeRF to be learned from monocular videos. Neural3DVideo21 \cite{li2022neural} also transformed NeRF into a space-time domain, and achieved frame-interpolation by interpolating time latent vectors. However, the time-consuming rendering process of above NeRF-style methods limit their capabilities to real-time applications. Directly warping images to novel views according to depth is an efficient view-synthesis pipeline. \textit{DynSyn20}~\cite{yoon2020novel} combined multi-view and single-view depths to generate temporal consistent depths for dynamic views warping.  However, their method has two drawbacks: first, it requires foreground masks that separate static and dynamic contents; second, their method can not handle occlusions well. \textit{3DMaskVol21}~\cite{lin2021deep} proposed a method of generating dynamic MPI with a 3D mask volume to alleviate artifacts around the integration boundary of background and instantaneous MPIs. However, their method requires two-step training and background images. Thus, limiting their general capabilities. Compared with NeRF-style methods, \textit{DynSyn20} and \textit{3DMaskVol21}, our method is efficient on rendering and compact on storage.

\noindent\textbf{Neural learnable basis.}
Our method is closely related to basis learning~\cite{liu2012robust}. In signal processing, data often contains underlying structure that can be processed intelligently by linear combinations of \textit{subspaces}. Tang~et al.~\cite{tang2020lsm} learned subspace minimization for low-level vision tasks, such as interactive segmentation, video segmentation, optical flow estimation and stereo matching. PCA-Flow\cite{wulff2015efficient} predicted video's optical flows as a weighted sum of the basis flow fields. We take inspiration from these works, and learn coefficients to combine globally shared time-wise subspace to draw instantaneous MPIs.

\begin{figure*}{
\centering
\includegraphics[width=0.8\linewidth]{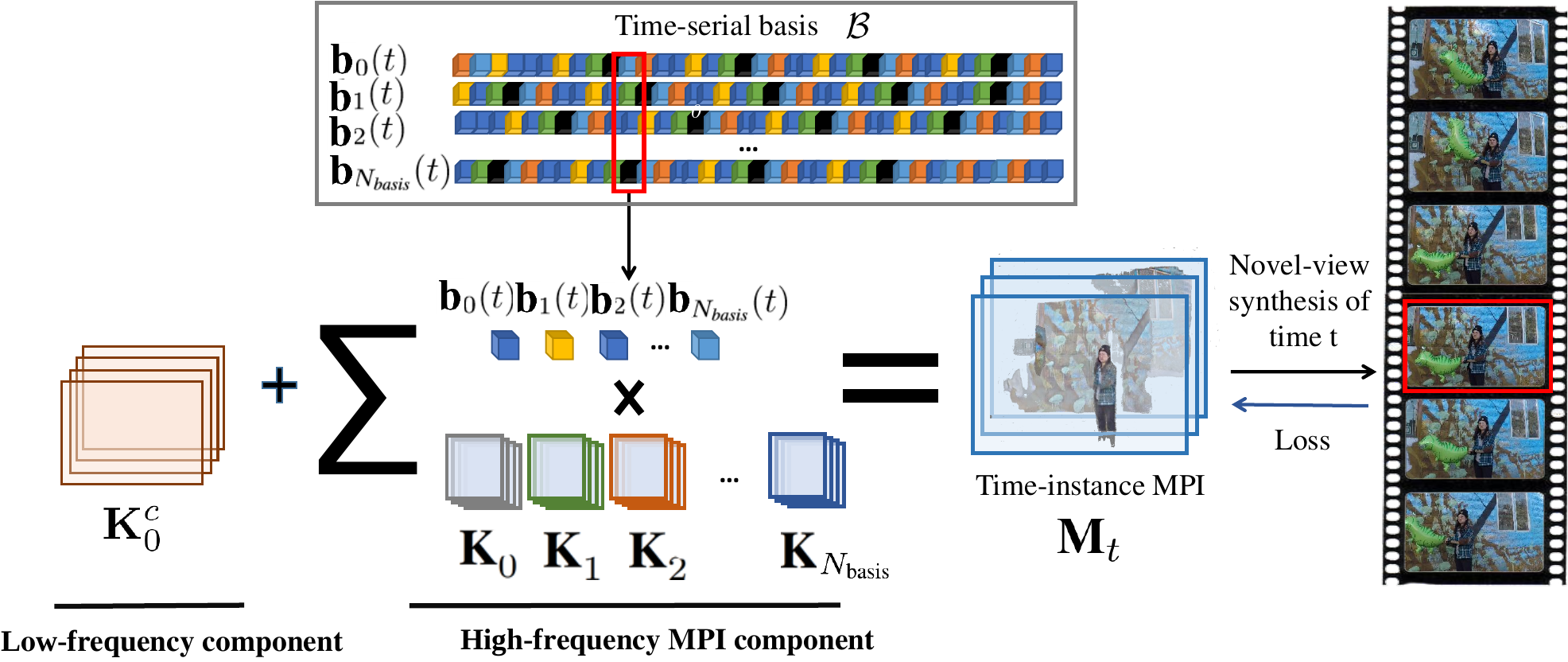}
\caption{Overall pipeline. The proposed Temporal-MPI contains three parts: low-frequency component $\mathbf{K}_0^c$, temporal basis $\mathcal{B}$ and high-frequency coefficients $\{\mathbf{K}_{n}\}_{n=1}^{N_\text{basis}}$. The alpha and color values in time-instantaneous MPI $\mathbf{M}_{t}$ are recovered as linear combinations of bases $\mathcal{B}$ and high-frequency coefficients $\{\mathbf{K}_{n}\}_{n=1}^{N_\text{basis}}$, and adding low-frequency component $\mathbf{K}_0^c$ from Temporal-MPI $\hat{\mathcal{M}}$. The color in the corresponding frame is rendered from time-instantaneous MPI $\mathbf{M}_{t}$ as MPI's alpha composition in Equation~(\ref{eq:alpha_compo}). The overall pipeline is differentiable and optimized per scene by pixel rendering loss. }
\label{fig:everall_pipe}
}\end{figure*}

\section{Approach}
Given a set of synchronized multi-view videos $\{\mathbf{I}_t^k\}$ of a dynamic scene, where $t=1,2,.\cdot\cdot\cdot T$ are the frame number, and $k=1,2,\cdot\cdot\cdot,K$ are camera indices, our goal is to construct a \textit{compact} 3D representation which enables \textit{real-time} and \textit{novel-view} synthesis of the dynamic contents at a given time $t\in[1,T]$.
To achieve the goal, one naive option is to calculate and save a set of separate MPI $\mathcal{M}=\{\mathbf{M}_{t} \in \mathbb{R}^{H \times W \times D \times 4}\}_{t=1}^{T}$ for every video frame. This, however, will be extremely memory- and computation-inefficient (generating $\mathcal{M}$ incurs more than 225$\times T$ MB data and around 2 seconds delay when rendering at VGA resolution \cite{lin2021deep}). As such, rather than having to calculate and save MPIs for all video frames in advance or having to calculate an MPI on-the-run, we investigate a novel Temporal-MPI representation with learned temporal basis to compactly encode high-frequency variation throughout the entire video. An overall pipeline of our approach is shown in Fig.~\ref{fig:everall_pipe}.

In the following, we will first briefly introduce the vanilla MPI representation in Section \ref{sec:mpi}. Then, the temporal basis formulation will be elaborated in Section \ref{sec:formulation}, and the novel-view temporal reconstruction in Section \ref{sec:reconstruction}.

\subsection{The Multi-plane Image Representation} \label{sec:mpi}
Being one of the seminal representation frameworks for 3D content embedding and novel-view synthesis, Multi-plane Images (MPI) learn a layered depth decomposition of the scene from a set of multi-view references \cite{zhou2018stereo,mildenhall2019llff,zhou2018stereo}. Following the MPI's illustration in Nex \cite{Wizadwongsa2021NeX},
let $D$ denote the number of depth layers in a MPI, with the dimension of each layer being $H\times W \times 4$, where $H$ and $W$ denote the height and width of the MPI layers, 4 denotes 3-channel RGB and 1-channel alpha $\alpha$. So we denote an MPI representation as $\mathbf{M}=\{\mathbf{C}_d,\mathbf{A}_d\}_{d=1}^{D}$, where $\mathbf{C}_{d} \in \mathbb{R}^{H\times W\times3}$ are multiple layers of 3-channel RGB images and $\mathbf{A}_{d} \in \mathbb{R}^{H\times W\times1}$ are one-channel alpha images, $d$ denotes the depth plane index.

Synthesizing novel-views $\mathbf{\hat{I}}$ based on the MPI $\mathbf{M}$ involves two steps: first, warp all depth planes in the MPI homographically from a reference view to a source view; and second, render pixels using alpha-composition \cite{10.1145/964965.808606} over each layer's color:
\begin{align}\label{eq:alpha_compo}
\hat{\mathbf{I}} = \mathcal{O}(\mathcal{W}(\mathbf{A}),\mathcal{W}(\mathbf{C})),
\end{align}
here $\mathcal{W}$ denotes the warping operator, and $\mathcal{O}$ denotes the compositing operator. The compositing operator $\mathcal{O}$ is defined as:
\begin{align}\label{eq:alpha+weighting}
\mathcal{O}(\mathbf{A},\mathbf{C}) = \sum_{d=1}^{D}\mathbf{C}_{d}\mathcal{T}_d(\mathbf{A}),\\ \mathcal{T}_{d}(\mathbf{A}) = \mathbf{A}_{d}\prod_{i=d+1}^{D}(1-\mathbf{A}_{i}).
\end{align} where $\prod_{i=d+1}^{D}(1-\mathbf{A}_{i})$ are accumulated transmittance, $\mathcal{T}_{d}$ are opacity. The output of $\mathcal{O}(\mathbf{A},\mathbf{C})$ are final rendered colors.
Both the composition $\mathcal{O}$ and the warping $\mathcal{W}$ operations are differentiable, thus allowing the representation $\mathbf{M}$ to learn the geometry and color information from final pixel rendering loss.

\subsection{Temporal Basis Formulation} \label{sec:formulation}

At a given time instance $t\in [1, T]$, we denote the time-instance MPI as $\mathbf{M}_{t}$. In order to render the entire novel view sequence at sequential timestamps, a set of time-instance MPIs $\mathcal{M}=\{\mathbf{M}_{t} \in \mathbb{R}^{H \times W \times D \times 4}\}_{t=1}^{T}$ are needed to generate. Based on the afore-analyzed reasons, we cannot exhaustively calculate and save $\mathcal{M}$. We propose a novel Temporal-MPI representation which is able to encode the rich 3D and high-frequency variation information throughout the entire video as compact temporal basis, and in the meantime, preserve high rendering efficiency for real-time novel-view synthesis.
To achieve this, we divide the goal into two tasks, i.e., (i) learning the low-frequency color components as explicit parameters, and (ii) learning the high-frequency variation over a set of temporal basis.

\subsubsection{Explicit Parameter Learning for Low-frequency Component}

Low-frequency contents in a video constitute the low-frequency part of the total energy along the time dimension, which can be well-captured and modeled explicitly by time-invariant parameters.
By treating all the frames of the multi-view video $\{\mathbf{I}_t^k\}_{t,k}$ as source views equally and ignoring their respective frame indices, we can directly learn the multi-plane time-invariant RGB color parameters $\mathbf{K}^\text{c}_{0} \in \mathbb{R}^{ H\times W\times D/8\times  3}$ using the pixel rendering loss. $\mathbf{K}^\text{c}_{0}$ models the low-frequency energy of the video, with possible blur over the dynamic area. Such an explicit modelling scheme for the low-frequency component proves to be important \cite{Wizadwongsa2021NeX}  and let the subsequent dynamic modelling to better focus on the temporal variation.

\subsubsection{Temporal Basis Learning for High-Frequency Contents}

Compared with low-frequency components, the high-frequency contents in $\mathcal{M}$ constitute the high-frequency energy along the time dimension. Being high-dimensional and with dynamic variations, the high-frequency contents still constitute a highly regularized manifold, considering the fact that (i) the video length is limited (we model video with 24 frames in length, although these frames could be extracted from longer video sequences), and (ii) time-variant pixels within a scene usually show consistent motion in clusters. This motivates us to compactly represent the high-frequency components based on a few learned time-variant temporal basis.

We denote the temporal basis as $\mathcal{B}\in\mathbb{R}^{4\times T \times N_\text{basis}}$ which span the temporal variation space for $\mathcal{M}$. Here $N_\text{basis}$ denotes the total number of basis. The first dimension of $\mathcal{B}$ is set to 4 which is reserved for modelling both the MPI color component (with 3 channels): $\mathcal{B}^{c}=\{\mathbf{b}^c_n\}_{n=1}^{N_\text{basis}}$, and the alpha component $\mathcal{B}^{\alpha}=\{\mathbf{b}^{\alpha}_n\}_{n=1}^{N_\text{basis}}$ (1 channel). Therefore $\mathcal{B}=[\mathcal{B}^{c},\mathcal{B}^{\alpha}]$.

In our proposed framework, the temporal basis will be estimated by two time-dependent functions which are Multi-Layer Perceptron (MLP) networks $\mathcal{V}^c$ and $\mathcal{V}^\alpha$:
\begin{align}
\{\mathbf{b}^c_n(t)\}_{n=1}^{N_\text{basis}}= \mathcal{V}^c(\mathcal{E}(t)):~\mathbb{R} \mapsto \mathbb{R}^{3\times N_\text{basis}},\\
\{\mathbf{b}^\alpha_n(t)\}_{n=1}^{N_\text{basis}}=\mathcal{V}^\alpha(\mathcal{E}(t)):~\mathbb{R} \mapsto \mathbb{R}^{1\times N_\text{basis}}.
\end{align}
Here $\mathcal{E}(\cdot)$ is a time-encoding function which encodes time-sequential information into a high dimensional latent vector \cite{li2022neural}. The temporal basis $\mathcal{B}$ learns a parsimonious frame that efficiently spans the temporal variation manifold. With a \textit{pixel-specific} coding coefficient (to be elaborated in the next section), $\mathcal{B}$ can efficiently model the MPI pixel's temporal variation throughout the entire video.

\subsection{Temporal Coding for Novel-view Synthesis} \label{sec:reconstruction}
For an arbitrary frame index $t\in[1,T]$, a time-instance MPI $\mathbf{M}_t=[\mathbf{A}_t, \mathbf{C}_t]$ can be constructed based on the temporal basis $\mathcal{B}$ according to:
\begin{align}
\mathbf{C}_t(\mathbf{x})     &= \mathbf{K}_0^c(\mathbf{x}) + \sum_{n=1}^{N_\text{basis}}\mathbf{K}^c_n(\mathbf{x}) \times  \mathbf{b}_n^c(t), \label{eq:infer_color}\\
\mathbf{A}_{t}(\mathbf{x}) &= \sum_{n=1}^{N_\text{basis}}\mathbf{K}^{\alpha}_{n}(\mathbf{x}) \times \mathbf{b}_{n}^{\alpha}(t). \label{eq:infer_alpha}
\end{align}
Here $\mathbf{K}^{\alpha}_{n}(\mathbf{x})$ and $\mathbf{K}^c_n(\mathbf{x})$ are the coding coefficients for the respective temporal basis at a given MPI spatial location $\mathbf{x}\in\mathbb{R}^3$ (the 3 dimensions of $\mathbf{x}$ include its 2D coordinates and the depth plane index in $\mathcal{M}_t$). These coding coefficients are estimated by another set of MLPs $\mathcal{K}^c$ and $\mathcal{K}^\alpha$:
\begin{align}
\{\mathbf{K}^c_n(\mathbf{x})\}_{n=1}^{N_\text{basis}}&= \mathcal{K}^c(\mathcal{R}(\mathbf{x})): \mathbb{R}^{3}\mapsto \mathbb{R}^{3\times N_\text{basis}},\\
\{\mathbf{K}^\alpha_n(\mathbf{x})\}_{n=1}^{N_\text{basis}}&=\mathcal{K}^\alpha(\mathcal{R}(\mathbf{x})): \mathbb{R}^{3}\mapsto\mathbb{R}^{1\times N_\text{basis}}.
\end{align}
Similarly, here $\mathcal{R}(\cdot)$ is a position-encoding function which encodes the spatial information $\mathbf{x}$ into high-dimensional representations \cite{mildenhall2020nerf}.

Based on Equation (\ref{eq:infer_color}) and (\ref{eq:infer_alpha}),
the time-instance MPI $\mathbf{M}_t$ can be warped and composited to any arbitrary viewing angles according to Equation (\ref{eq:alpha+weighting}) and (\ref{eq:alpha_compo}). In addition, by querying all elements $t=1,\cdot\cdot\cdot, T$ along the temporal basis, we can construct the time-instance MPI for each video frame.

\textbf{Remarks.} (i) our proposed temporal MPI representation composes of an explicitly learned low-frequency multi-plane color component $\mathbf{K}^c_0 \in \mathbb{R}^{H \times W \times D/8 \times 3}$, and a dynamically coded time-variant component via simultaneous basis and coefficient learning. We have achieved compression along the temporal dimension via the temporal basis, which compactly encodes time-variant color and geometry variation information throughout the entire video.

(ii) To maintain rendering efficiency and save storage-space, spatial-temporal information is efficiently encoded and propagated among different components in the Temporal-MPI. First, the low-frequency component $\mathbf{K}_0^c$ is temporally shared among all time frames, this ensures overall reconstruction quality and enables the high-frequency components to focus on time-dependent variations only; and second, the high-frequency coefficients, i.e., $\{\mathbf{K}^c_n(\mathbf{x})\}_{n=1}^{N_\text{basis}}$ and $\{\mathbf{K}^\alpha_n(\mathbf{x})\}_{n=1}^{N_\text{basis}}$, are point-wisely coded/learned, however, over a common set of temporal basis. This helps to remove the redundancy in modelling dynamic variation, and also helps to remove motion ambiguities for some pixels.

\subsection{Training Loss Function}
 To let the Temporal-MPI focus on reconstruction quality, we ignore the sparsity of coding coefficients for this task. Coefficients and the temporal basis are jointly learned and optimized. The whole system is optimized via the following loss function $\mathcal{L}$:
 \begin{equation}
    \mathcal{L}=||\hat{\mathbf{I}}_t^k - \mathbf{I}_t^k||_2 + \lambda_1 ||\nabla\hat{\mathbf{I}}_t^k - \nabla \mathbf{I}_t^k ||_1 + \lambda_2 \text{TVC}(\mathbf{K}_0^c),
\end{equation}
where $\hat{\mathbf{I}}_t^k$ is the rendered image at time $t$ for the camera $k$, $\mathbf{I}_t^k$ is the ground truth image from the same view. The first term in $\mathcal{L}$ calculates the $L_2$ reconstruction loss. The second term penalises edge inconsistencies, with $\nabla$ denoting the gradient operator. In the third term, $\text{TVC}$ denotes total variation loss \cite{chambolle1997image}. $\lambda_1$ and $\lambda_2$ are balancing weights for different loss terms.

\begin{figure*}
    \centering

    \renewcommand{\arraystretch}{0.5}
    \begin{tabular}{ccccc}

            \includegraphics[width=1.8cm, clip]{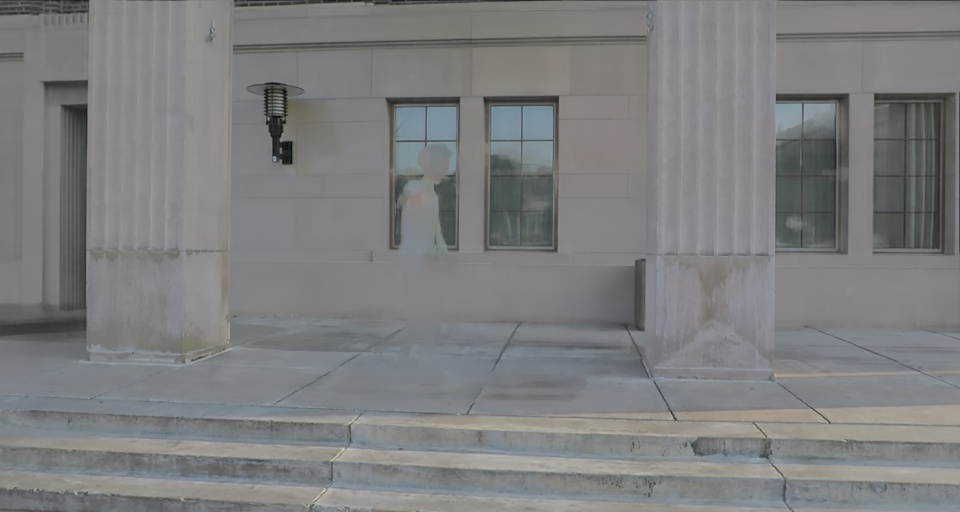}
        &
            \includegraphics[width=1.8cm, clip]{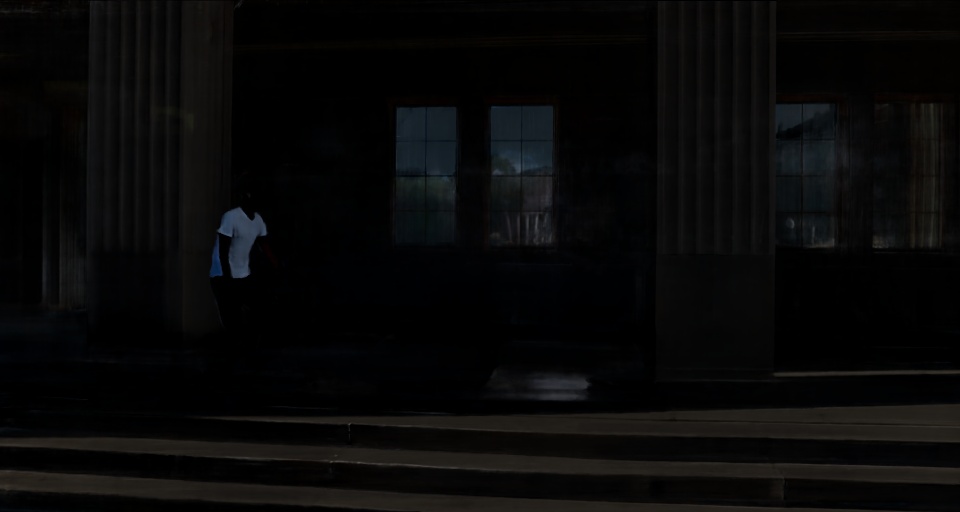}
        &

            \includegraphics[width=1.8cm, clip]{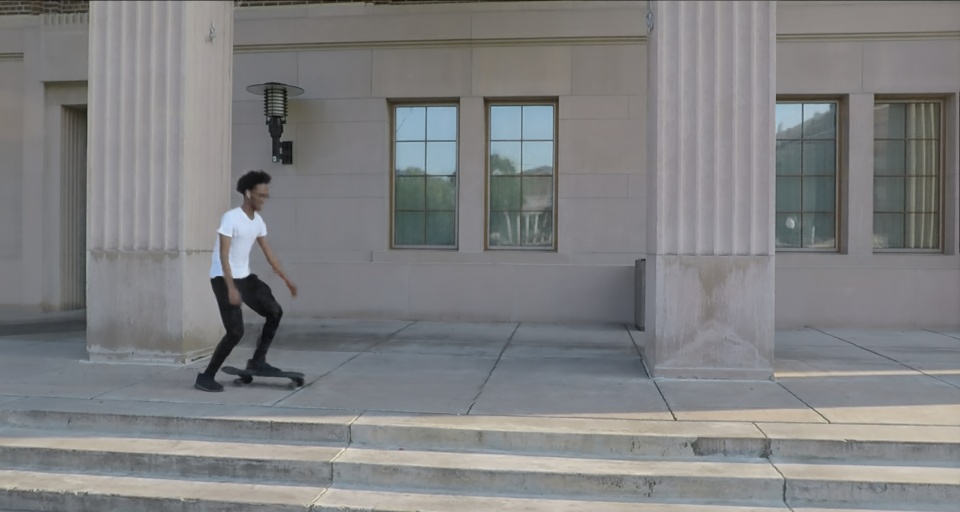}
        &

            \includegraphics[width=1.8cm, clip]{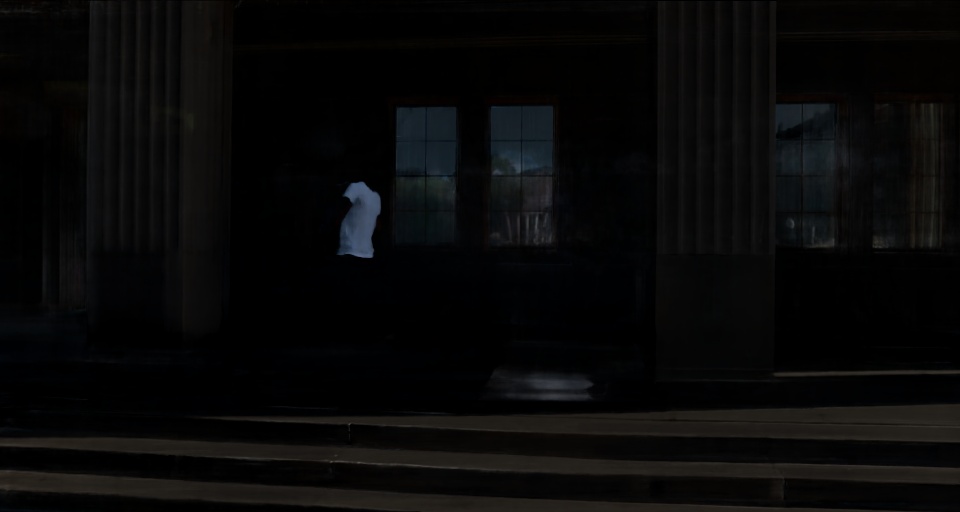}
        &

            \includegraphics[width=1.8cm, clip]{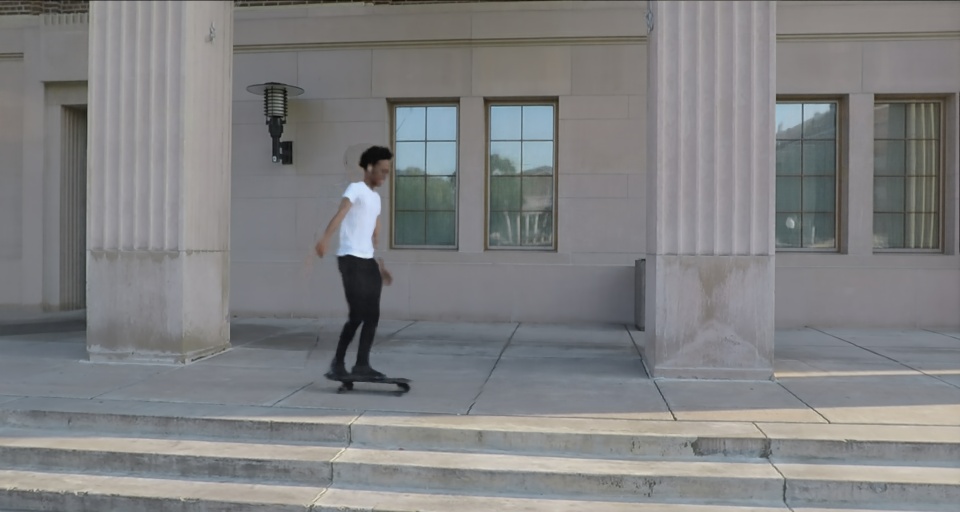}
        \\
                    \includegraphics[width=1.8cm, clip]{figure/skating_2/background.jpg}

        &
            \includegraphics[width=1.8cm, clip]{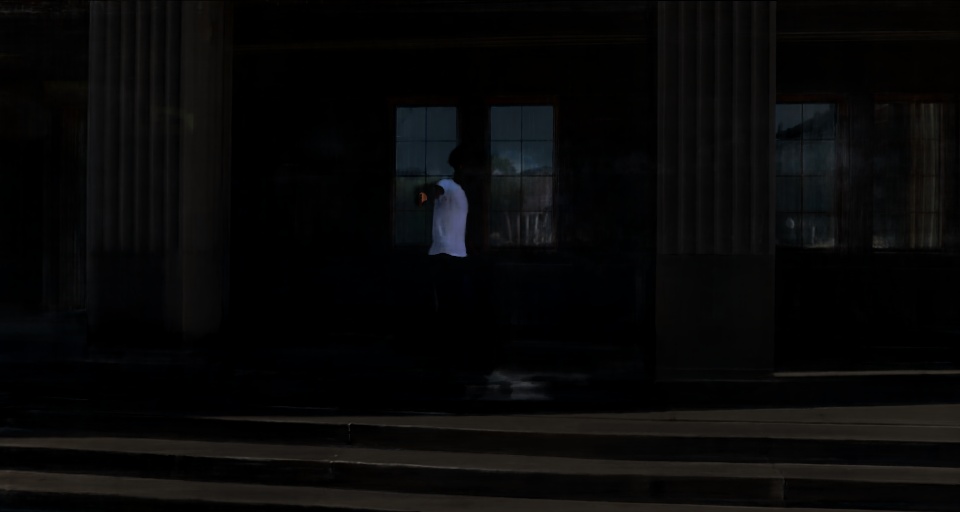}
        &
            \includegraphics[width=1.8cm, clip]{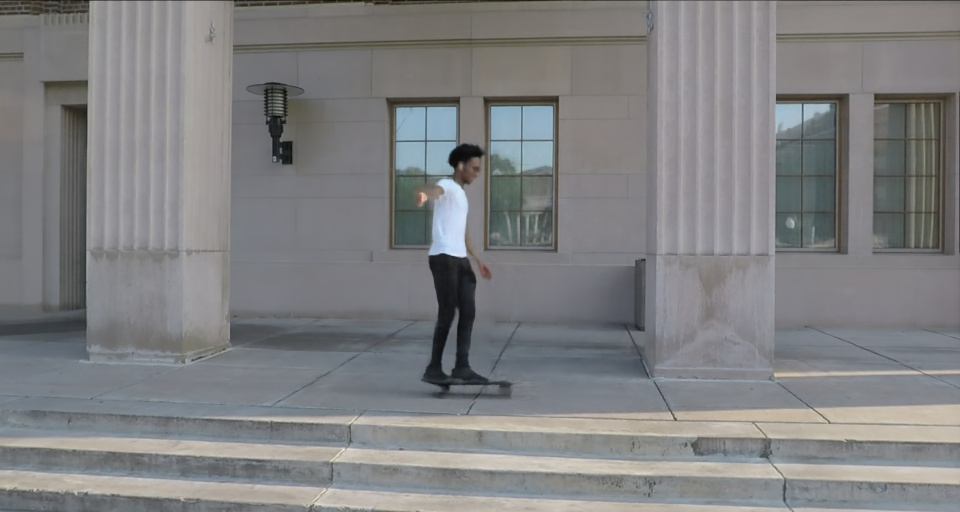}
        &
            \includegraphics[width=1.8cm, clip]{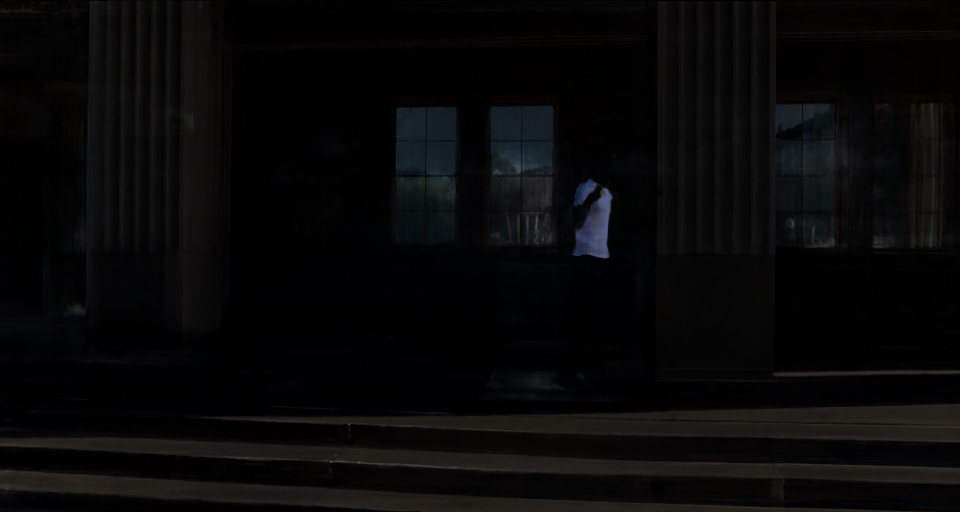}
        &
            \includegraphics[width=1.8cm, clip]{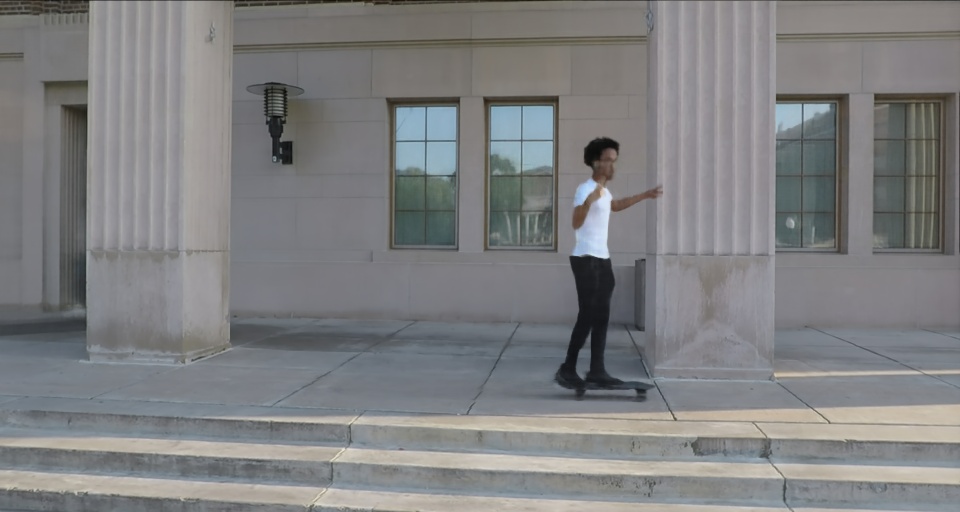}

        \\

        \includegraphics[width=1.8cm, clip]{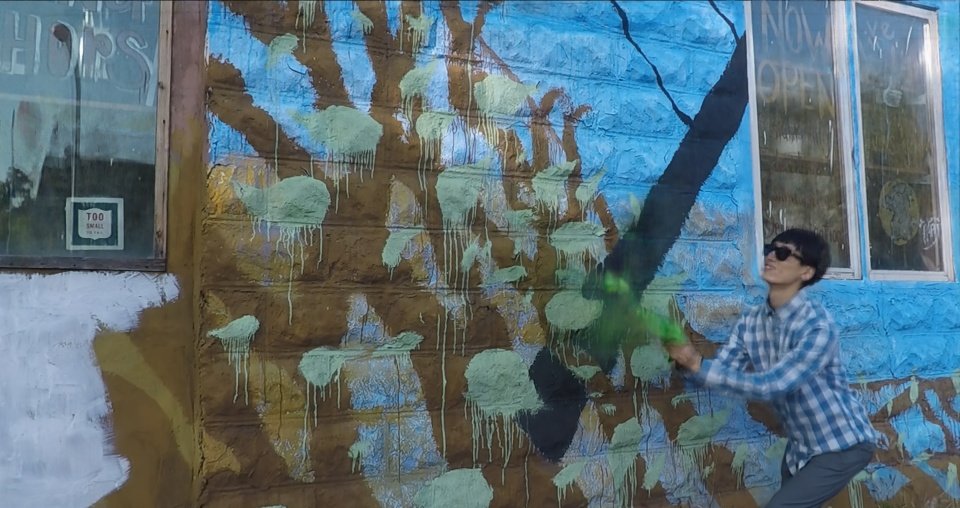}

        &

            \includegraphics[width=1.8cm, clip]{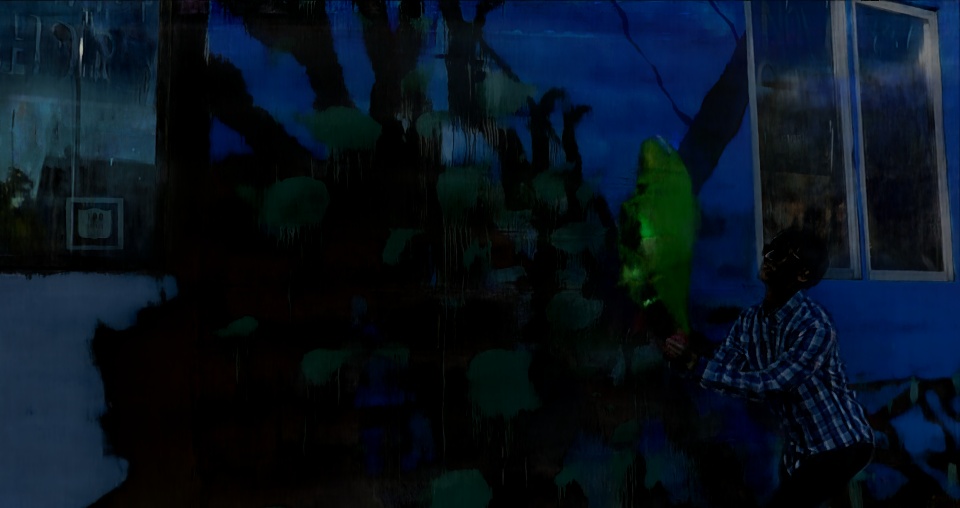}
        &

            \includegraphics[width=1.8cm, clip]{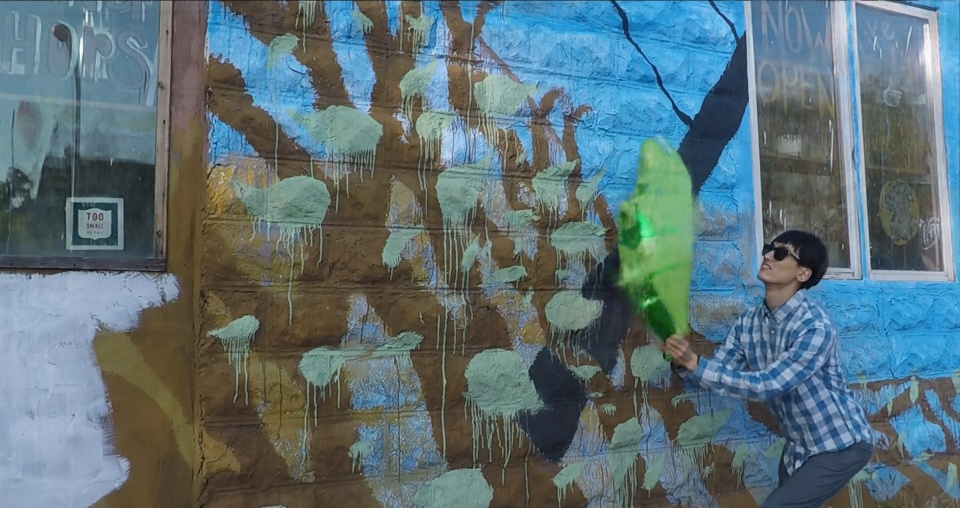}
        &

            \includegraphics[width=1.8cm, clip]{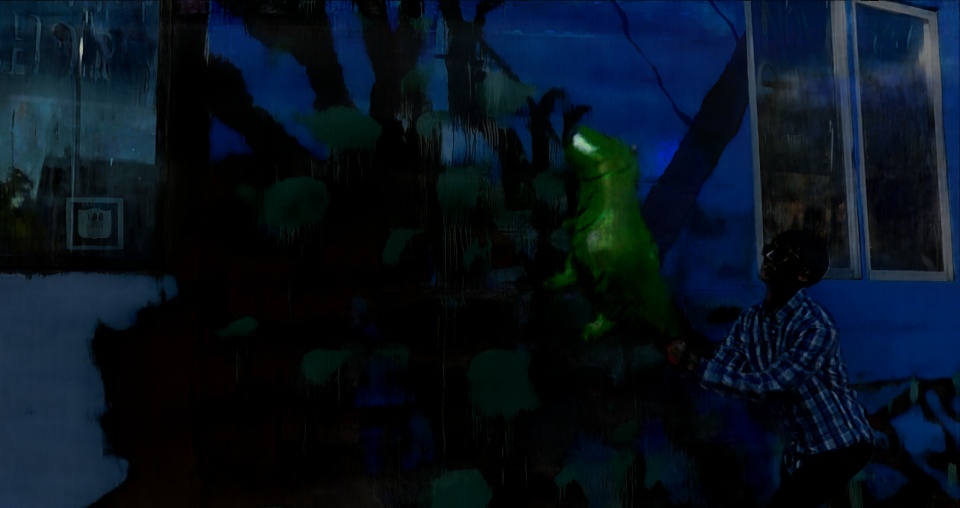}
        &

            \includegraphics[width=1.8cm, clip]{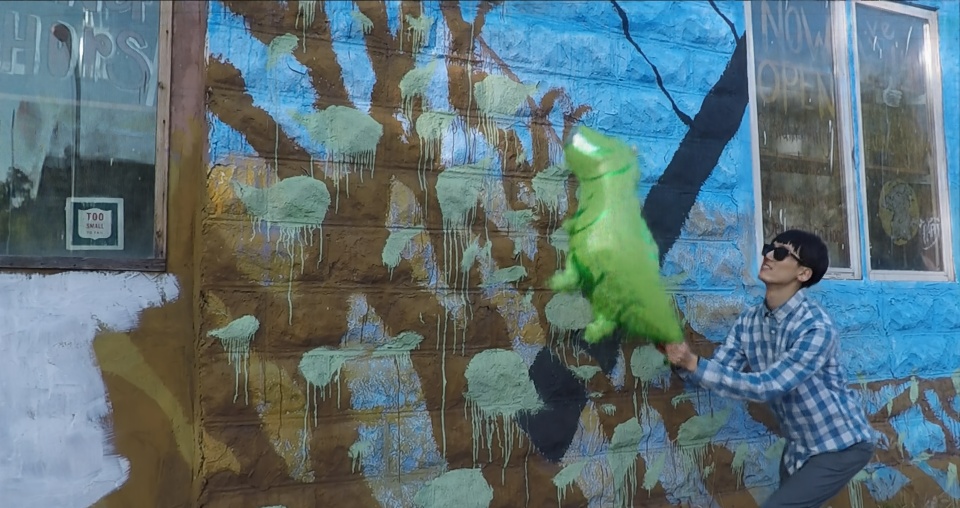}
        \\

        \includegraphics[width=1.8cm, clip]{figure/balloon_2-2/background.jpg}

        &
            \includegraphics[width=1.8cm, clip]{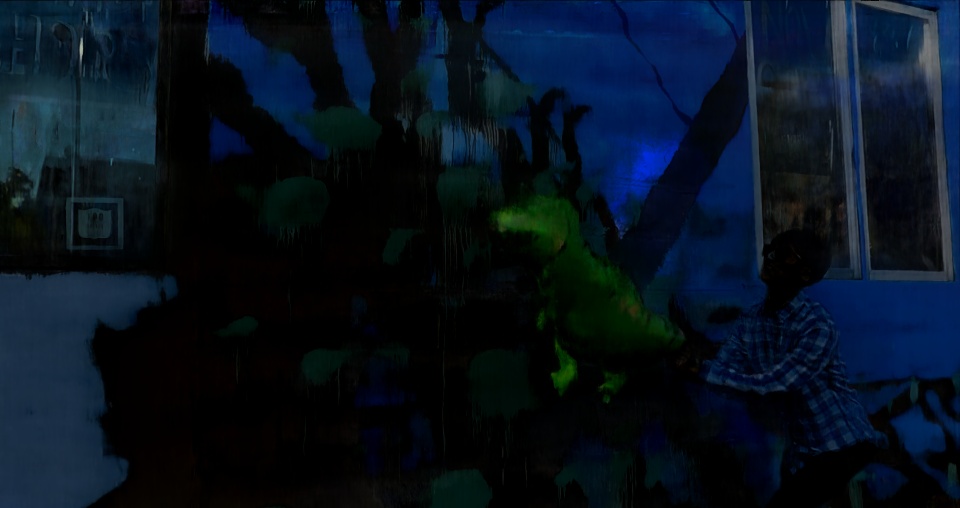}
        &
            \includegraphics[width=1.8cm, clip]{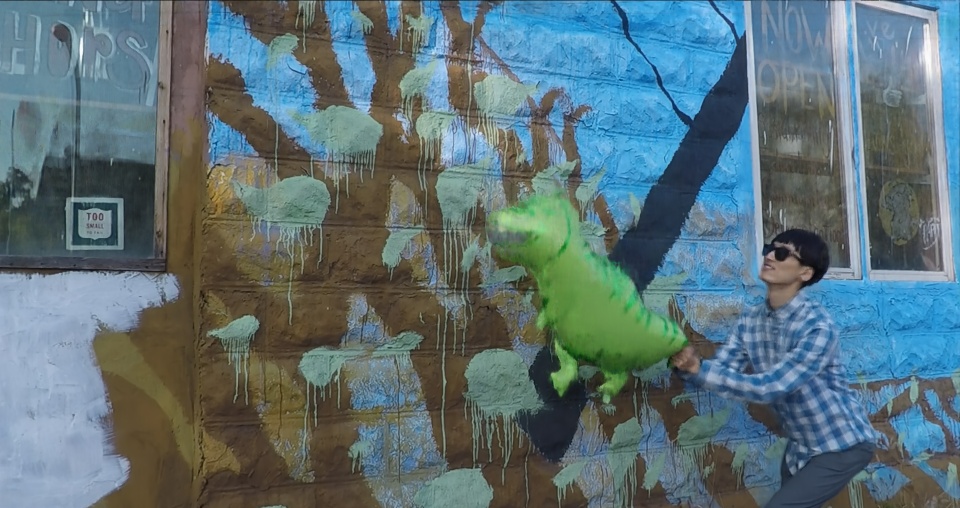}
        &
            \includegraphics[width=1.8cm, clip]{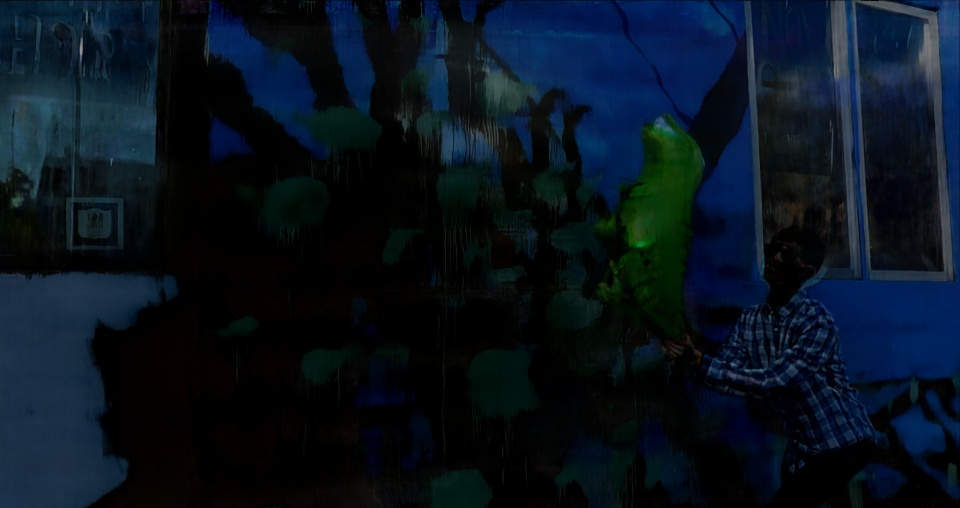}
        &
            \includegraphics[width=1.8cm, clip]{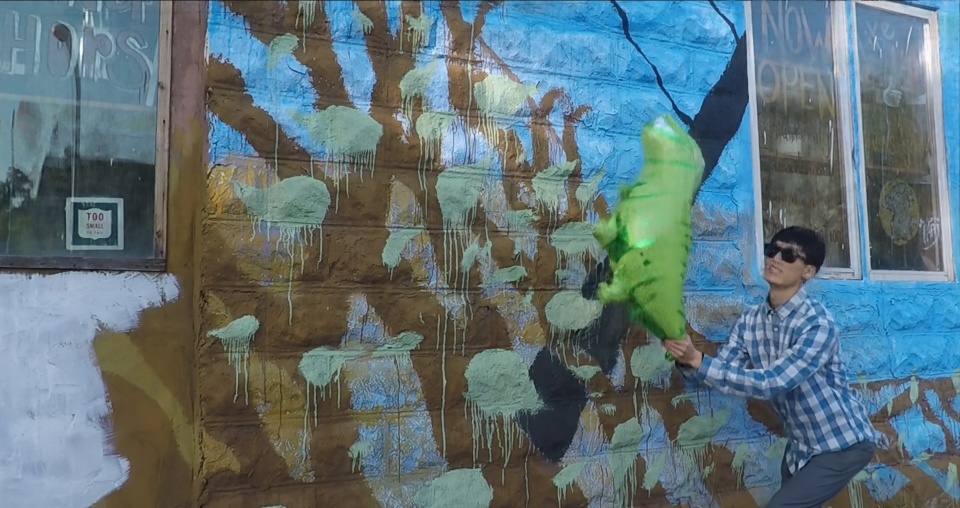}

                 \\

            \includegraphics[width=1.8cm, clip]{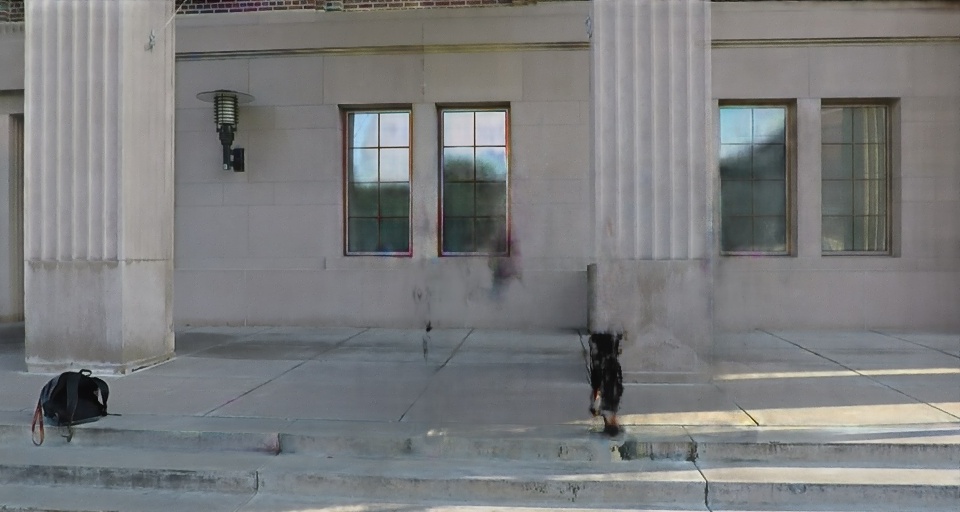}

        &

            \includegraphics[width=1.8cm, clip]{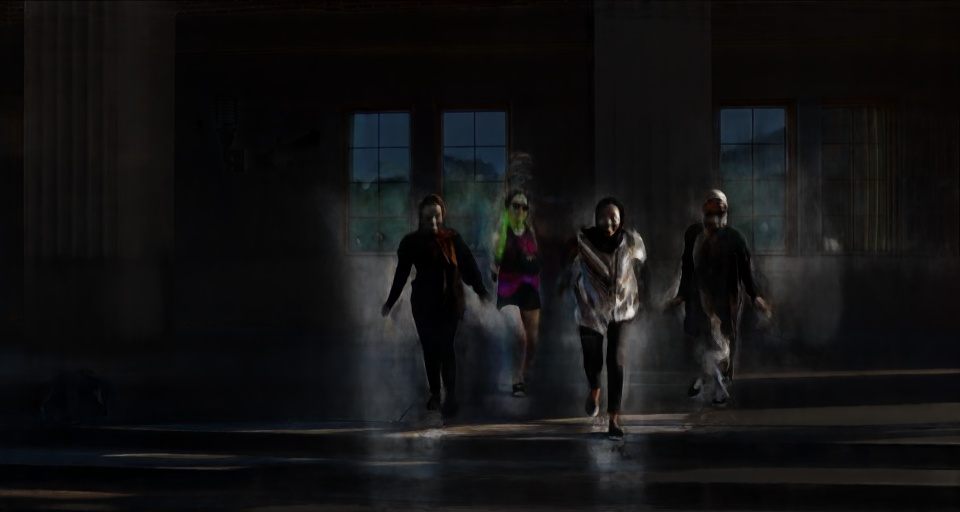}
        &

            \includegraphics[width=1.8cm, clip]{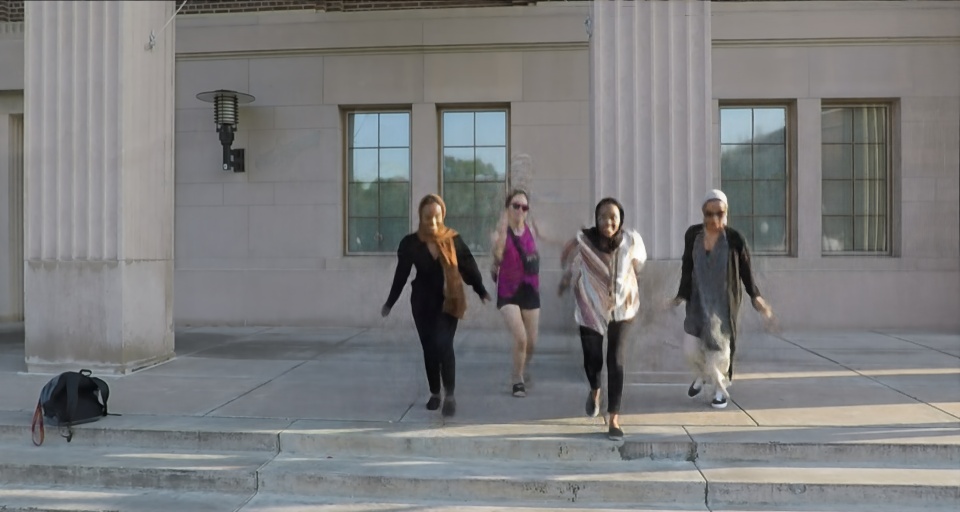}
        &

            \includegraphics[width=1.8cm, clip]{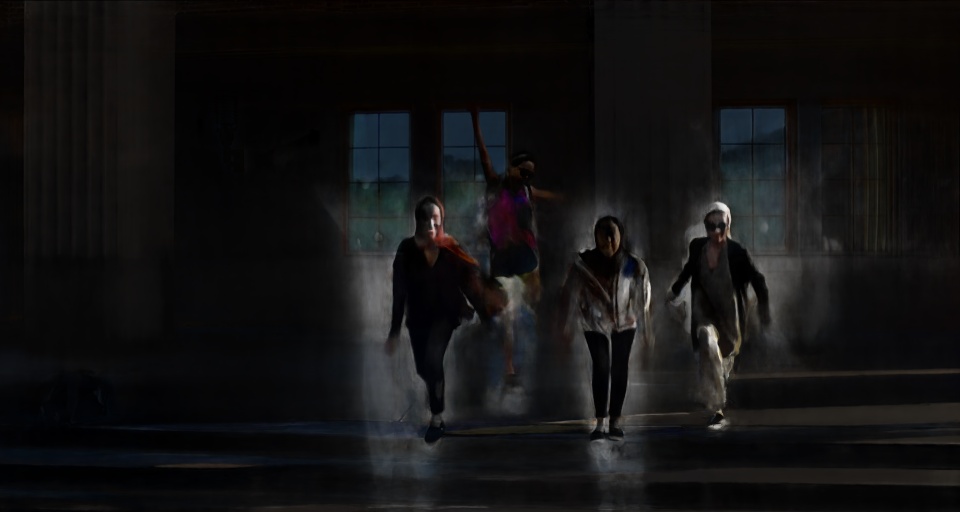}
        &

            \includegraphics[width=1.8cm, clip]{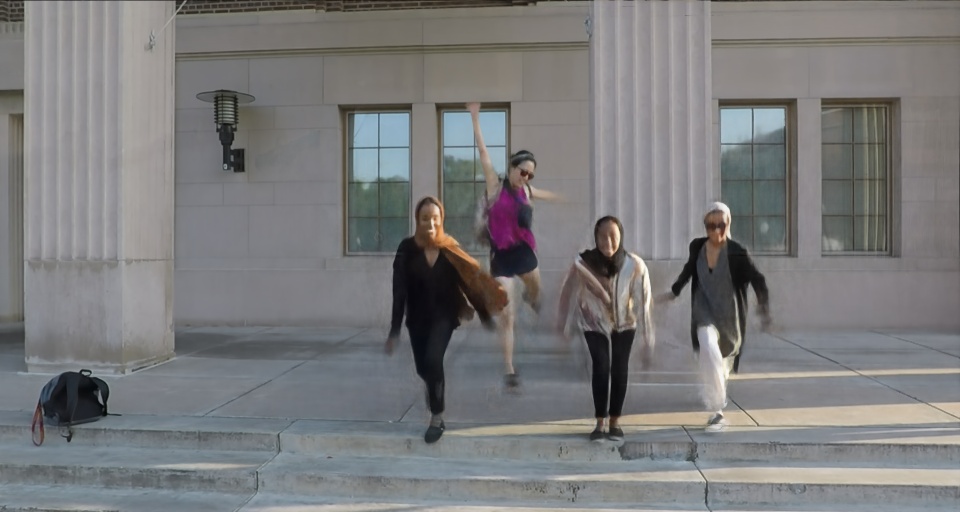}
        \\

            \includegraphics[width=1.8cm, clip]{figure/jumping/background.jpg}

        &
            \includegraphics[width=1.8cm, clip]{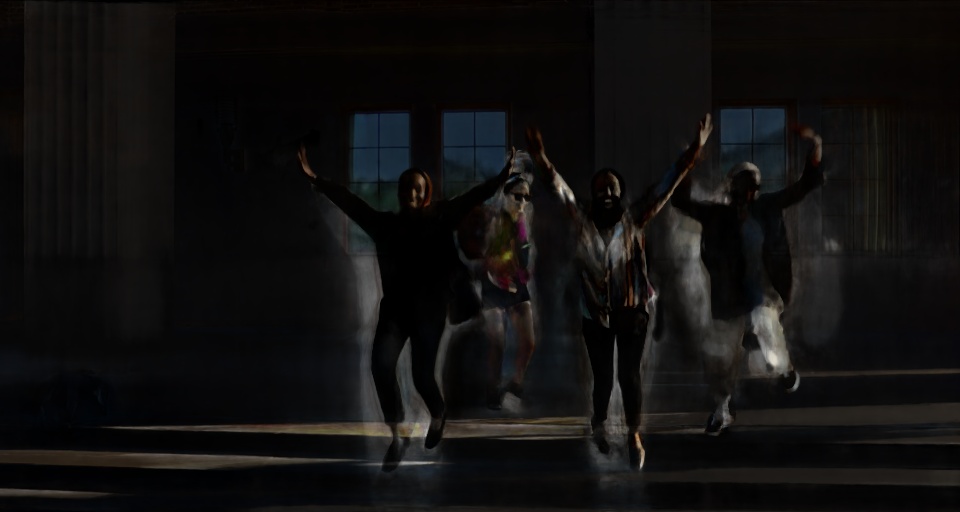}
        &
            \includegraphics[width=1.8cm, clip]{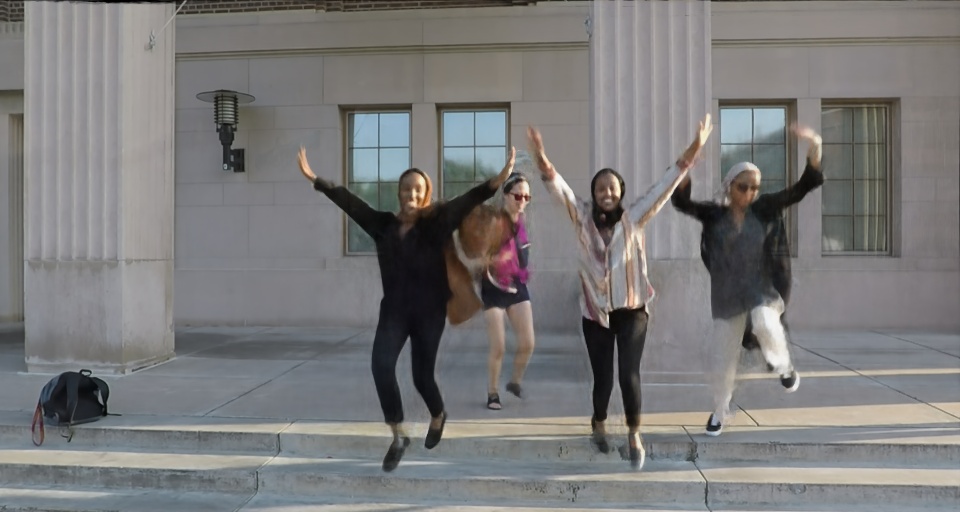}
        &
            \includegraphics[width=1.8cm, clip]{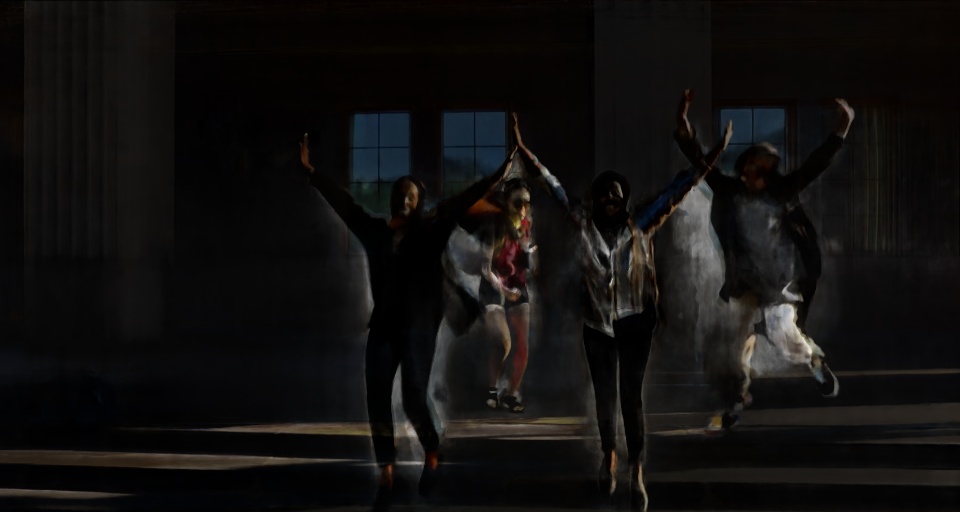}
        &
            \includegraphics[width=1.8cm, clip]{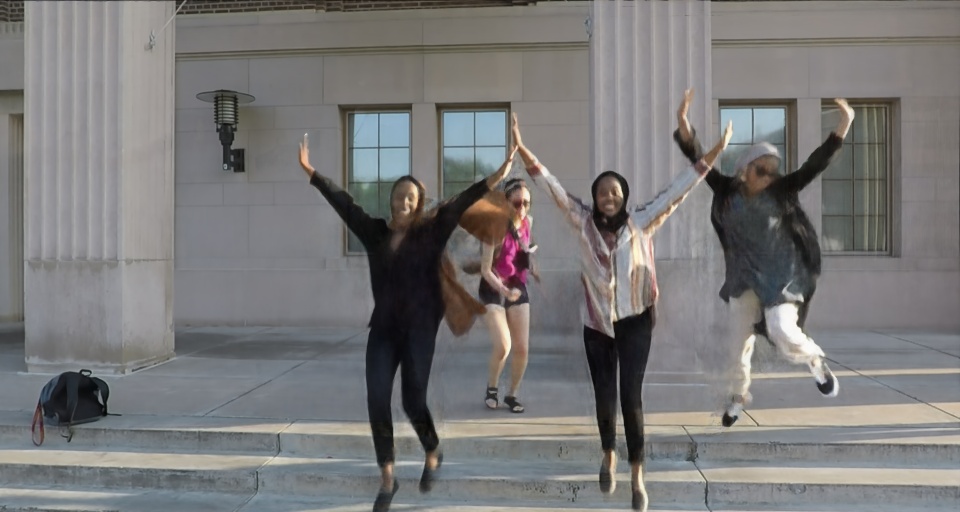}
\\
            \vphantom\scriptsize (a) Low-frequency
        &
            \vphantom\scriptsize (b) w/o
        &
            \vphantom\scriptsize (c) Full
        &
            \vphantom\scriptsize (d) w/o
        &
            \vphantom\scriptsize (e) Full

    \end{tabular}
    \caption{Low-frequency scene representation ablation study. The low-frequency components are rendered in (a); color output from the high-frequency components are rendered in (b) and (d); full rendering are visualised in (c) and (e).}
  \label{fig:global_ablation}
\end{figure*}

\section{Experiments}

\subsection{Implementation Details}

Our model is implemented in PyTorch 1.10, using Adam as optimiser. The initial learning rate is set as 0.001, and decay by 0.1 every 2000 steps. The model takes 16 hours to be trained on one Nvidia Geforce RTX 2070 Super GPU with a batch of 1500 rays, using 5.3 GB memory.
The output resolution is 576 $\times$ 300. The position-encoding method in \cite{mildenhall2020nerf} is formulated as $\mathcal{R}(p)=[sin(2^{0}\frac{\pi}{2}p),sin(2^{0}\frac{\pi}{2}p)$,
$\dots,sin(2^{l}\frac{\pi}{2}p),cos(2^{l}\frac{\pi}{2}p)]$ where the input location of scene point is normalised to [-1, 1] and $l$ is the index of encoding level set as 3. The index of time is embedded into a latent vector in size of 32 using dictionary learning as in \cite{li2022neural}. For networks that parameterise $\mathcal{K}^c$ and $\mathcal{K}^{\alpha}$, we use MLP networks with 8 layers and 384 hidden nodes. Networks for $\mathcal{V}^{c}$ and $\mathcal{V}^{\alpha}$ are using MLP with 4 layers and 64 hidden nodes. The shape of high-frequency coefficients $\{\mathbf{K}^c_n(\mathbf{x})\}_{n=1}^{N_\text{basis}}$ and $\{\mathbf{K}^\alpha_n(\mathbf{x})\}_{n=1}^{N_\text{basis}}$ in Temporal-MPI is $320 \times 596 \times 32 \times 4 \times 5$ where 32 is the number of planes $D$, 596 and 320 are width $W$ and height $H$ including marginal offsets set as 10, 4 includes 3 channels of colors and 1 channel of alpha, and 5 is the number of basis $N_\text{basis}$. The shape of temporal basis $\mathcal{B}$ is $4\times 5 \times 24$ where 5 is the number of basis $N_\text{basis}$, 24 is the total number of timestamps and 4 includes 3 channels for color and 1 channel for alpha. Low-frequency component $\mathbf{K}_{0}^{c}$ is in the shape of $ 320 \times 596 \times 4 \times 3$ before the repetition along depth dimension.

\subsection{Dataset}

\begin{figure}[h]
    \centering
    \includegraphics[width=0.4\linewidth]{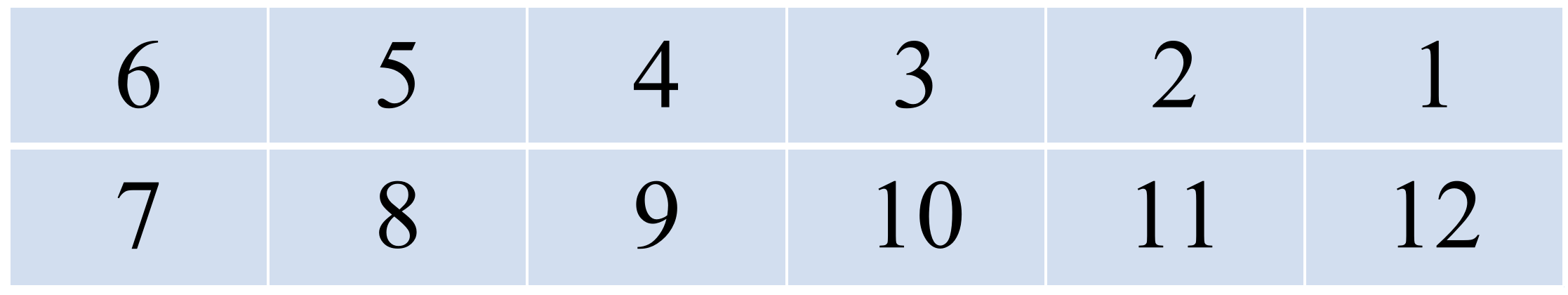}
    \caption{Camera indexes in camera array.}
    \label{fig:camera_array}
\end{figure}

Our model is trained and evaluated on the Nvidia Dynamic Scenes Dataset \cite{yoon2020novel} that contains 8 scenes with motions recorded by 12 synchronized cameras. Nvidia Dynamic Scenes Dataset captures a dynamic scene with static background via stationary cameras which suit our goal of separately learning low- and high-frequency components well. We extract camera parameters for every camera using COLMAP \cite{schoenberger2016sfm}. We extracted 24 frames from the video sequence, and used multi-view images in selected frames for training. We select camera views 1-11 for training, and camera 12 for testing. So the total number of training images is 264. The camera location arrangement is shown in Fig. \ref{fig:camera_array}.

\begin{figure*}[h]
    \centering
    \includegraphics[width=0.7\linewidth]{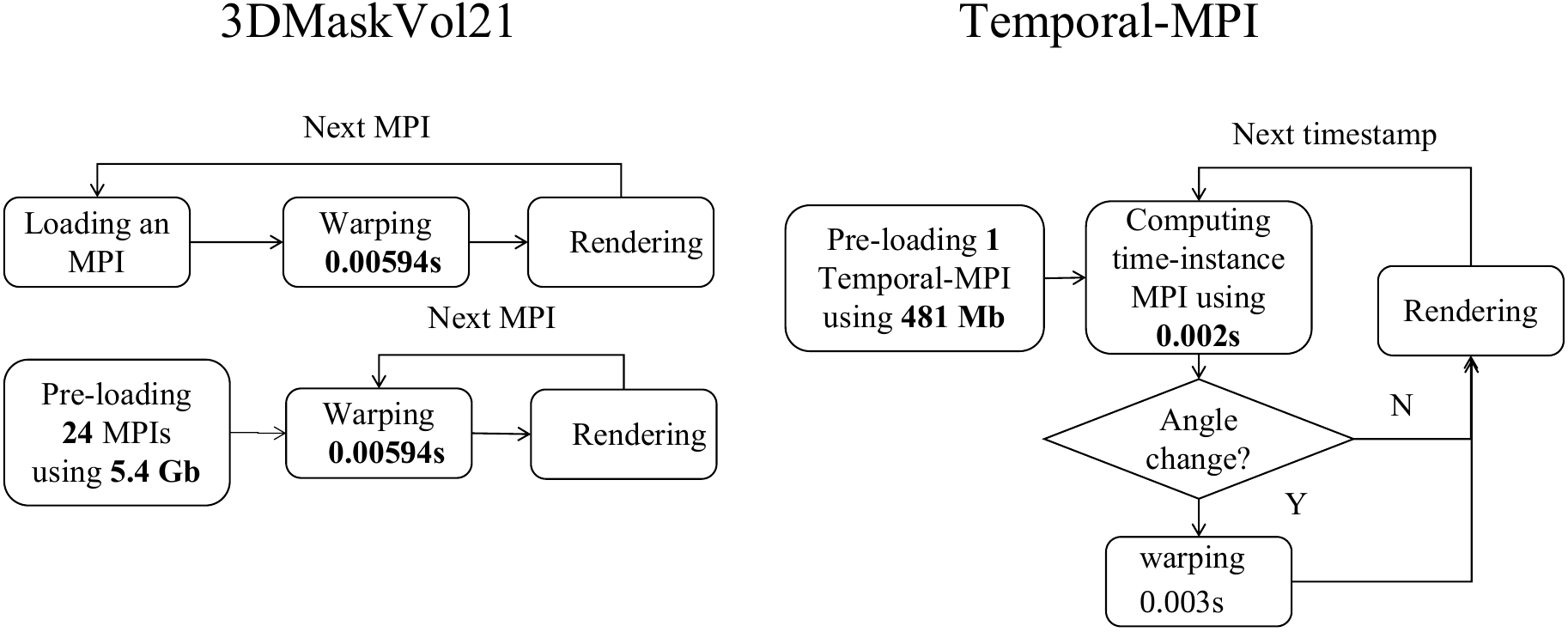}
    \caption{Rendering pipeline comparison. }
    \label{fig:flow}
\end{figure*}

\subsection{Ablation Study}
In this section, we investigate the effectiveness of our main contributions in Temporal-MPI: high-frequency coefficients, temporal basis and low-frequency component.

\subsubsection{Video Length}
 We evaluate the performance of our model trained with different length of videos. As shown in Table \ref{tab:psnr_video_frame}, We found performance degradation when the total number of timestamps $T$ increased. This is due to reaching the representation threshold of temporal basis and high-frequency coefficients.

\begin{table}
\centering
\caption{Ablation study on PSNR vs. total number of timestamps. With  $D$ as 32 Planes.}
\begin{center}\label{tab:psnr_video_frame}
\renewcommand\tabcolsep{3.5pt}
\begin{tabular}{lllllllll}
  \toprule
  \multirow{2}{*}{Scene/PSNR}  & \multicolumn{6}{c}{Number of timestamps} \\
            & 8  & 16  & 24 &32 &40 & 48 &60 \\ \midrule
Skating-2  &30.323 &	28.813 &28.575	& 28.612 & 27.324&	25.431	&25.012 \\
Truck-2   &28.441 & 28.174&28.056 &27.951 & 27.591&25.332 & 24.963\\
Jumping   &25.850 &25.661 &25.486 & 25.301&25.001 & 24.759 & 23.854\\
Balloon2-2   & 25.775&25.381 &25.171 & 24.893&24.557 &24.474 & 23.945\\
  \bottomrule
\end{tabular}
\end{center}
\bigskip\centering
\label{tab12}
\end{table}

\subsubsection{Low-frequency Colors and High-frequency Coefficients}
To validate the contributions of low-frequency component $\mathbf{K}_0^c$, high-frequency components and time-serial basis ($\{\mathbf{K}^c_n(\mathbf{x})\}_{n=1}^{N_\text{basis}}$, $\{\mathbf{K}^\alpha_n(\mathbf{x})\}_{n=1}^{N_\text{basis}}$, $\mathcal{B}$), we performed experiments without these modules. As shown in Table \ref{tab:ablation_study},  without low-frequency component or high-frequency components will lead to a worse result than the full model. We separately render the low-frequency and high-frequency parts of the Temporal-MPI to prove their individual contributions. The visualization of separate rendering settings are shown in Fig. \ref{fig:global_ablation}. The low-frequency component in Fig. \ref{fig:global_ablation} is calculated by directly summing $\mathbf{K}_0^c$ across depth planes. We wish to highlight that it is designed to facilitate the MLP to focus on modelling the high-frequency residual by explicitly modelling the low frequency content. It can be seen that the low-frequency components successfully capture the low-frequency energy of the video, while the high-frequency components complement the low-frequency ones to produce high quality rendering for dynamic scenes.

\begin{table}
\centering
\caption{Ablation study of low-frequency and high-frequency components.}
\begin{center}\label{tab:ablation_study}
\renewcommand\tabcolsep{3.5pt}
\begin{tabular}{lllll}
  \toprule
  \multirow{2}{*}{Methods} &\multirow{2}{*}{No. Planes} & \multicolumn{3}{c}{Metrics} \\
        &   & SSIM ($\uparrow$) & PSNR ($\uparrow$) & LPIPS ($\downarrow$) \\ \midrule
w/o low-frequency & 32 &0.192 & 10.3 & 0.726\\
  w/o high-frequency  & 32 & 0.611& 22.6& 0.213\\
 \midrule
 Full   & 32 & 0.859 &24.87 &0.196 \\
  \bottomrule
\end{tabular}
\end{center}
\label{tab_wo}
\end{table}

\begin{table}
\centering
\caption{Inference time vs. shape of Temporal-MPI.}
\begin{center}\label{tab:inference_time_ablation}
\renewcommand\tabcolsep{3.5pt}
\centering
\small{
\begin{tabular}{cccc}
  \toprule
  Resolution  & No.Basis  & No.Planes  & Inference Time (seconds, $\downarrow$) \\ \midrule
596$\times$320 & 5 & 32& 0.002\\
596$\times$320 & 13 & 32& 0.003\\
1038$\times$1940 & 5 & 32 & 0.025\\
1038$\times$1940 & 13 & 32 & 0.029\\
1038$\times$1940 & 5 & 192 & 0.030\\
  \bottomrule
\end{tabular}}
\end{center}
\label{tab:size_inference_time}
\end{table}

\begin{table*}[h]
\caption{Quantitative evaluation of novel view synthesis on the Dynamic Scenes dataset. MV denotes whether the approach uses multi-view information or not, Ind. Src denotes the index of source views used to train the model. Ours (D=32) denotes using 32 planes in MPI. \label{exp:nvidia_quan}}
\centering
\begin{tabular}{lclccccc}
\toprule
\multirow{2}{*}{Methods} & \multirow{2}{*}{Ind. Src} & \multirow{2}{*}{MV}   & \multicolumn{3}{c}{Metrics} \\
& & &  SSIM ($\uparrow$) & PSNR ($\uparrow$) & LPIPS ($\downarrow$) &  \\
\midrule
\textit{SynSin20}~\cite{wiles2020synsin}& $3$ & No  & 0.488 &	16.21 &	0.295\\
\textit{MPIs20}~\cite{single_view_mpi}&$3$   & No & 0.629 & 19.46	& 0.367 \\
\textit{3D Ken Burn19}~\cite{10.1145/3355089.3356528}  & $3$ & No & 0.630 & 19.25 &	0.185 \\
\textit{3D Photo20}~\cite{Shih3DP20} & $3$ & No  & 0.614 & 19.29	& 0.215 \\
\textit{NeRF20}~\cite{mildenhall2020nerf} &$1-11$  & Yes  & 0.893 & 24.90	& 0.098 \\
\textit{ConsisVideoDepth20}~\cite{10.1145/3386569.3392377} & $3$& Yes & 0.746 & 21.37 & 0.141 \\
\textit{DynSyn20}~\cite{yoon2020novel} & $1-11$ & Yes & 0.761 & 21.78 & 0.127 \\
\textit{NeuralFlow21}~\cite{li2021neuralsceneflow} & $3$ & Yes & \textbf{0.928} & \textbf{28.19} & \textbf{0.045} \\
\textit{D-NeRF21}~\cite{pumarola2021d} &$1-11$ & Yes& 0.334 & 17.05 & 0.545\\
\textit{3DMaskVol21}~\cite{lin2021deep} &$3,9$  & Yes & 0.603 & 20.10 & 0.285\\
\midrule
\textbf{Ours} (D=32) & $1-11$ & Yes   &0.859 &24.87&0.196\\
\bottomrule
\end{tabular}
\end{table*}

\subsubsection{Inference Speed}

In this section, we investigate the relationships between inference speed and the size of Temporal-MPI. From Table \ref{tab:size_inference_time}, we can find that the computations of linear combinations of basis are efficient, and a big volume size of Temporal-MPI will not affect its real-time performance. Inference speed experiments are conducted on one Nvidia Tesla V100 GPU.

\subsection{Evaluation and Comparison}

To prove the efficiency and compactness of our method, we first compare our method with state-of-the-art algorithms, \textit{3DMaskVol21}~\cite{lin2021deep} and \textit{NeuralFlow21}~\cite{li2021neuralsceneflow}, in terms of storage space and inference speed. Then, we evaluate the view synthesis quality with other methods.

\subsubsection{Evaluation on Compactness of Representation}
One of the main objectives of our approach is to learn a compact representation of a dynamic scene. So we evaluate the compactness of Temporal-MPI by comparing the number of network parameters and storage space with these two methods. As shown in Table \ref{tab:size_inference_time_2}, modeling a dynamic scene with 24 timestamps, Temporal-MPI only occupies 481 Mb for storage, which is 11 times smaller than \textit{3DMaskVol21} on storage space. So our approach is extremely fast and compact for real-time rendering.

\subsubsection{Evaluation on Efficiency}
From Table \ref{tab:size_inference_time_2},
we can find that i) rendering NeuralFlow21 requires querying MLP exhaustively, so it is the slowest on rendering. ii) our rendering time is much faster than 3DMaskVol21: as shown in Fig. \ref{fig:flow}, 3DMaskVol21 requires per-frame warping, which is not a mandatory step of ours; it also requires per-frame MPI loading, but we only need to load once for the entire sequence, therefore longer sequence will bring more advantages to our efficiency. Considering both loading and rendering time, ours are much faster on rendering than both NeuralFlow21 and 3DMaskVol21 (on a $T$=24 frame video). But 3DMaskVol21 is a generic method that generalizes to novel scenes, so it saves the cost of per-scene training. NeuralFlow21 has the highest rendering quality due to its advantages on dense sampling in depth dimensions.

\begin{table*}
 \centering
 \caption{Comparison on real-time rendering/inference speed, output resolution and storage space. We assume the length of the modeled dynamic video is 24 frames. \textit{3DMaskVol21} will require pre-loading 24 MPIs for real-time rendering of the whole sequence. \textit{NeuralFlow21} is impossible for real-time tasks. }
  \begin{tabular}{c c c c c}
    \toprule
    Time/Methods & \textit{NeuralFlow21} &\textit{3DMaskVol21}   & Ours\\
    \midrule
    MPI Generation Time (sec,$\downarrow$) & - & 2 & \textbf{0.002} \\
    MPI Loading Time (sec,$\downarrow$) & - & 0.043   & \textbf{0.083}/T \\
    Warping Time  (sec,$\downarrow$) & - & 0.00594   & \textbf{0.003} \\
    Rendering Time (sec/frame,$\downarrow$) & 6 & 0.049   & \textbf{0.008}(T=24) \\
    Output Resolution(pixel,$\uparrow$)  & 512 $\times$ 288 &  \textbf{640 $\times$ 360}  & 576$\times$ 300 \\
    Network Parameters (million, $\downarrow$)  & 5.26 & \textbf{1.17} & 6.00\\
    Storage Space (Mb,$\downarrow$) &--& 225$\times$T (24)=5400&  \textbf{481} $\times$ 1(D=32)\\
     Training Time (hour, $\downarrow$) &48$^\dag$   & 120   &  16$^\dag$   \\
    \bottomrule
    $\dag$ denotes scene specific training. \\
  \end{tabular}
  \label{tab:size_inference_time_2}

\end{table*}

\subsubsection{Evaluation on View-Synthesis Quality}
We evaluate the effectiveness of our approach by comparing it to baseline methods quantitatively and qualitatively. We compare our approach with state-of-the-art single-view or multi-view novel view synthesis methods. For monocular methods, we compare with \textit{SynSin20} \cite{wiles2020synsin} and \textit{MPIs20} \cite{single_view_mpi} trained on RealEstate 10K dataset \cite{zhou2018stereo}. \textit{3D Photo20} \cite{Shih3DP20} and \textit{3D Ken Burns19} \cite{10.1145/3355089.3356528} were trained by wild images. For multi-view methods, we compare with \textit{NeRF20} \cite{mildenhall2020nerf}, \textit{ConsisVideoDepth20} \cite{10.1145/3386569.3392377}, \textit{DynSyn20} \cite{yoon2020novel}, \textit{NeuralFlow21} \cite{li2021neuralsceneflow}, \textit{3DMaskVol21} \cite{lin2021deep} and \textit{D-NeRF21} \cite{pumarola2021d}. Results are referenced from recent publications \cite{li2021neuralsceneflow,lin2021deep}.
We document the rendering quality in three error metrics:  structural similarity index measure (SSIM), peak signal-to-noise ratio (PSNR), and perceptual similarity through LPIPS \cite{zhang2018unreasonable}. From Table \ref{exp:nvidia_quan}, our algorithm has competitive average score across three metrics.
Per-scene breakdown results are shown in Table \ref{tab:realffd}.

\begin{figure*}[t]
    \centering
    \setlength{\tabcolsep}{0.025cm}
    \renewcommand{\arraystretch}{0.5}
    \hspace*{-\tabcolsep}\begin{tabular}{cccccc}
    (a)
        &
            \includegraphics[width=3.0cm]{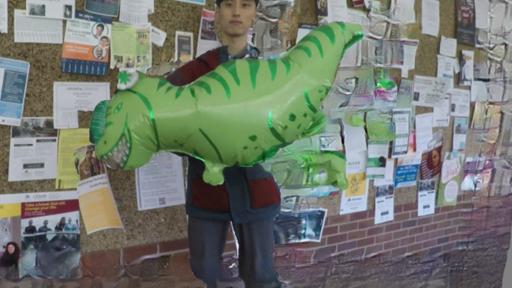}
        &
            \includegraphics[width=3.0cm]{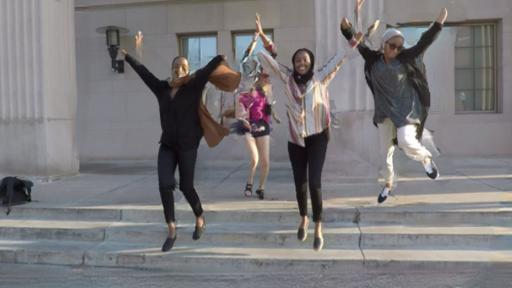}
        &
            \includegraphics[width=3.0cm]{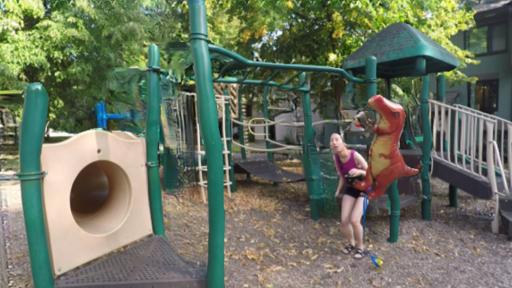}
        &
            \includegraphics[width=3.0cm]{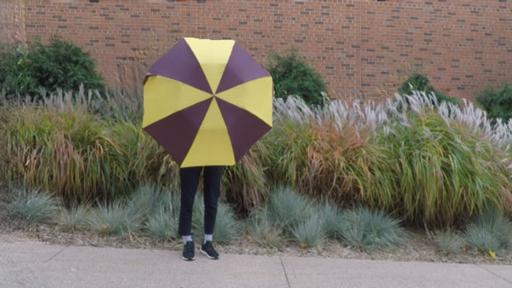}
        \\
        (b)
        &
            \includegraphics[width=3.0cm]{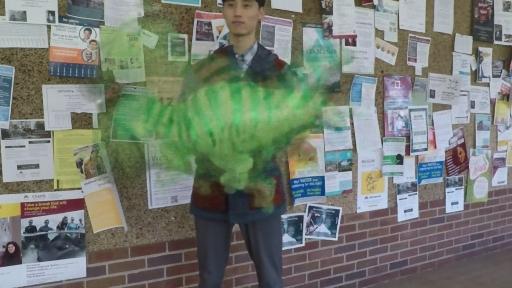}
        &
            \includegraphics[width=3.0cm]{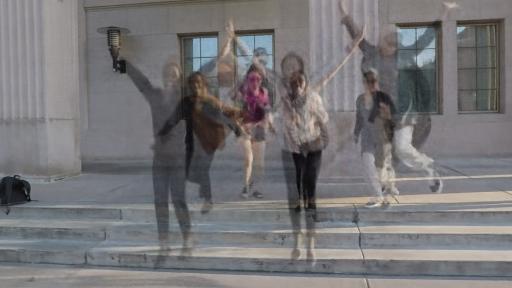}
        &
            \includegraphics[width=3.0cm]{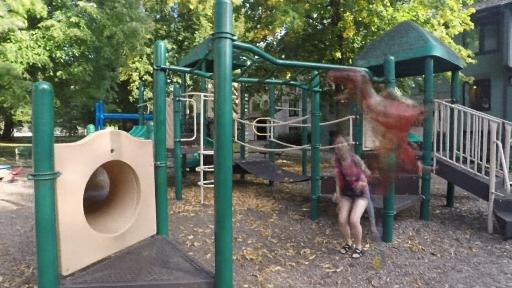}
        &
            \includegraphics[width=3.0cm]{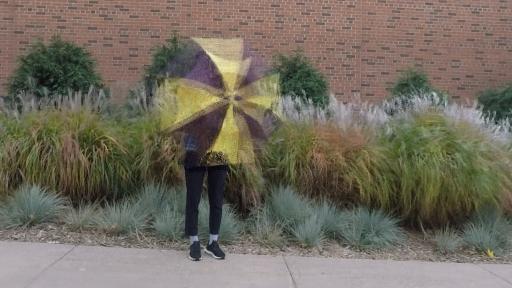}
               \\
          (c)
        &
            \includegraphics[width=3.0cm]{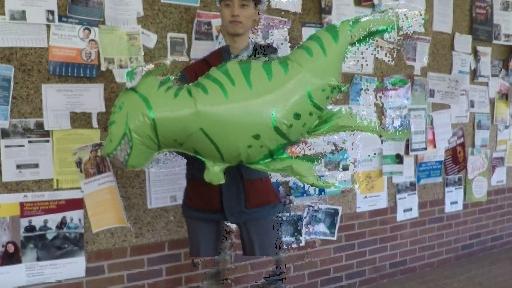}
        &
            \includegraphics[width=3.0cm]{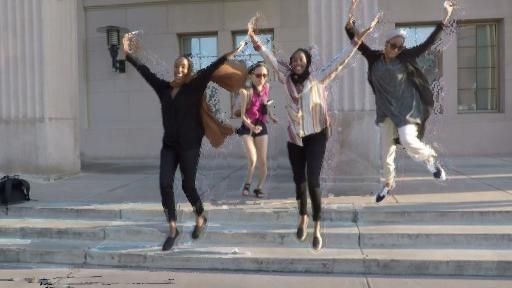}
        &
            \includegraphics[width=3.0cm]{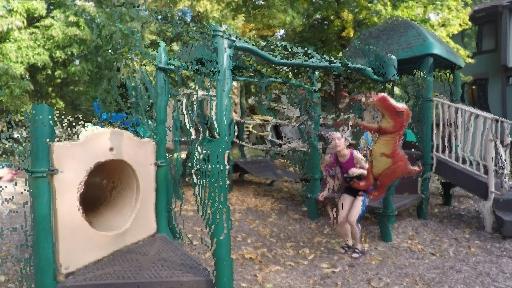}
        &
            \includegraphics[width=3.0cm]{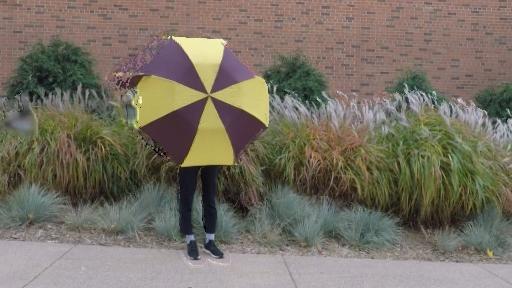}
       \\

        (d)
         &
                \includegraphics[width=3.0cm]{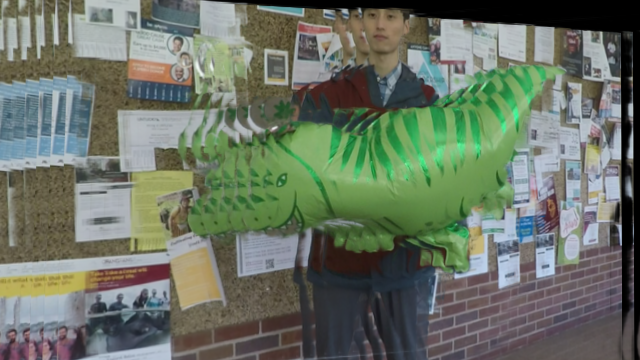}
        &
            \includegraphics[width=3.0cm]{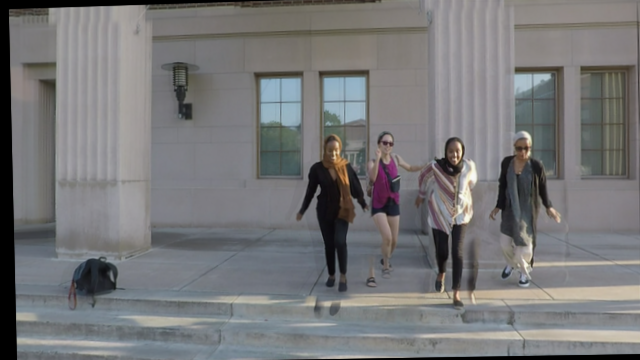}
        &
            \includegraphics[width=3.0cm]{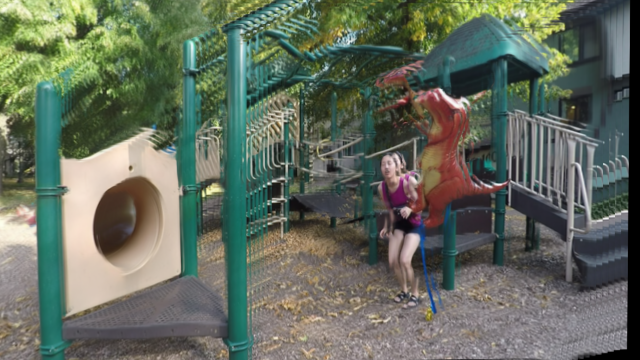}
        &
            \includegraphics[width=3.0cm]{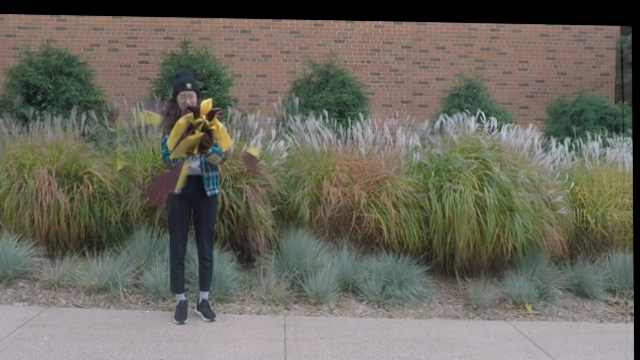}

                 \\

      (e)
     &
                \includegraphics[width=3.0cm]{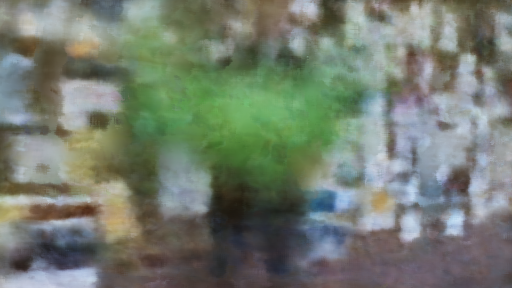}
        &
            \includegraphics[width=3.0cm]{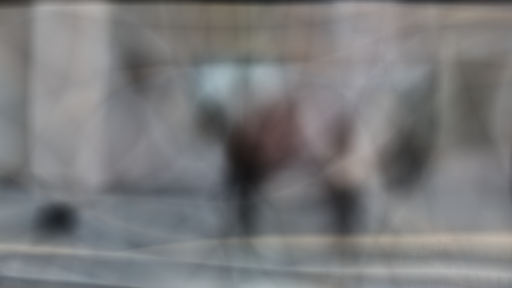}
        &
            \includegraphics[width=3.0cm]{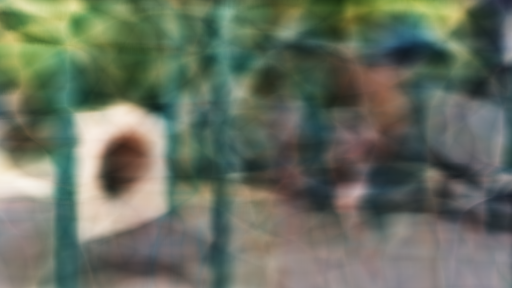}
        &
            \includegraphics[width=3.0cm]{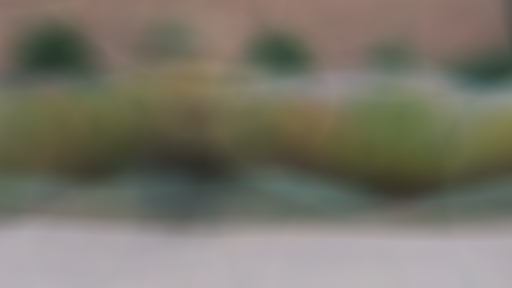}

                 \\

          \rotatebox{90}{\small{Ours}} &
                \includegraphics[width=3.0cm]{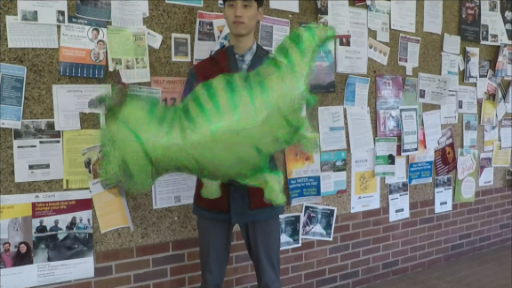}
        &
            \includegraphics[width=3.0cm]{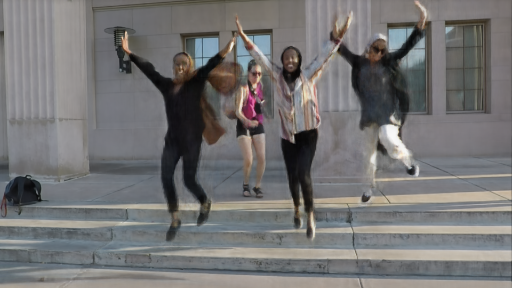}
        &
            \includegraphics[width=3.0cm]{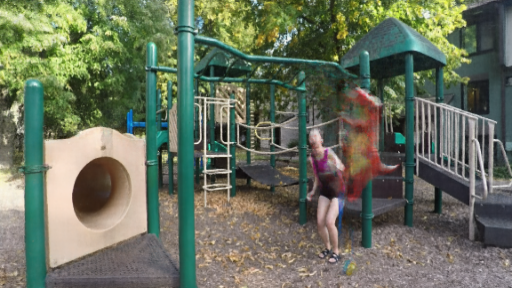}
        &
            \includegraphics[width=3.0cm]{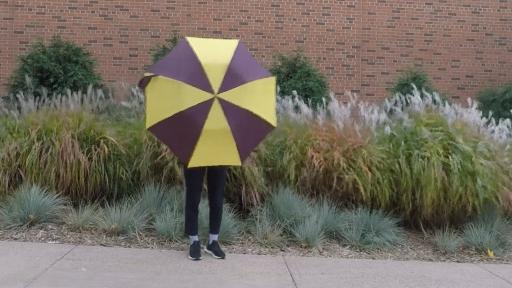}

                 \\
          \rotatebox{90}{\small{GT.}} &
            \includegraphics[width=3.0cm]{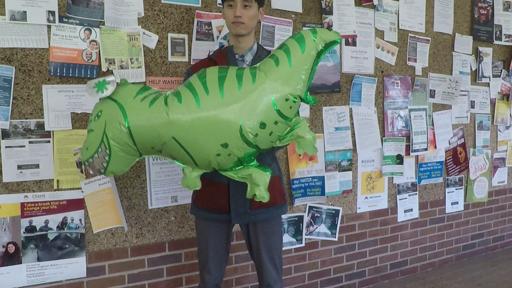}
        &
            \includegraphics[width=3.0cm]{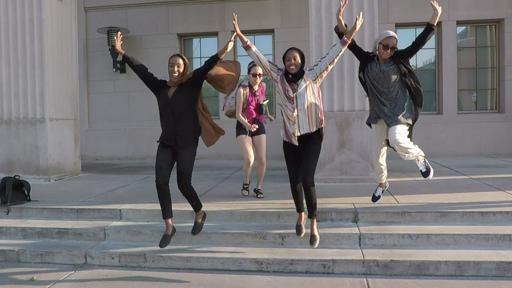}
        &
            \includegraphics[width=3.0cm]{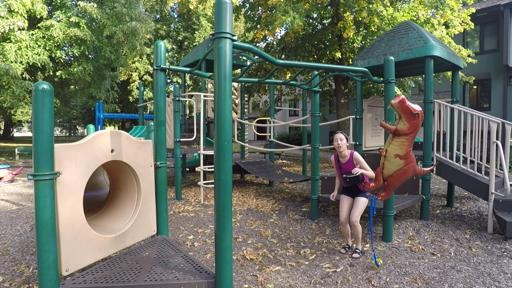}
        &
            \includegraphics[width=3.0cm]{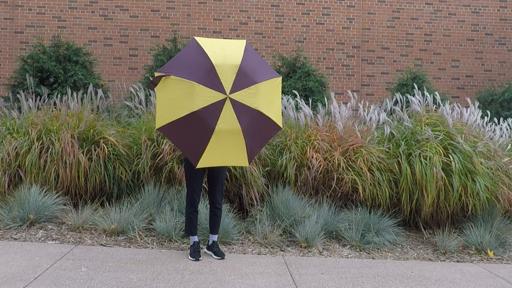}

        \\

    \end{tabular}
    \caption{Qualitative comparisons on the Dynamic Scenes dataset.
    }
    \label{fig:nvidia_qual}
\end{figure*}

Qualitative comparisons can be seen in Fig. \ref{fig:nvidia_qual}, which show that our method achieves competitive rendering quality in both low- and high-frequency parts. The visual results of \textit{3D Photo20} in Fig. \ref{fig:nvidia_qual} (a), \textit{NeRF20} in Fig. \ref{fig:nvidia_qual} (b) and \textit{DynSyn20} in Fig. \ref{fig:nvidia_qual} (c) are referenced from \cite{li2021neuralsceneflow}. Observed from above images, \textit{D-NeRF21} in Fig. \ref{fig:nvidia_qual} (e) produces blurry results, \textit{DynSyn20} has great artifacts on thin structures, \textit{3D Photo20} generates distortions, \textit{3DMaskVol21} produces ghosting effects around the object's boundary given scenes with forward moving motions, such as Jumping and Umbrella.

\begin{table*}
\centering
\caption{Per-scene breakdown results from \textit{DynSyn20}'s Dynamic Scenes dataset.}
\begin{center}\label{tab:realffd}
\begin{tabular}{lccccccc}
  \toprule
   & Skating-2 & Balloon1-2 &  Jumping &   Playground & Balloon2-2 & Truck-2 & Average\\
   \midrule
 {PSNR($\uparrow$)}  & 28.575 & 21.309 &25.486 & 20.594 & 25.171  &28.056 &24.865  \\
 {SSIM($\uparrow$)} & 0.925  & 0.802 &0.886& 0.7211 &0.885  & 0.937 &	0.859 \\
 {LPIPS($\downarrow$)} & 0.163 & 0.239  & 0.202 & 0.253& 0.171  & 0.150   & 0.196 \\
  \bottomrule
\end{tabular}
\end{center}
\bigskip\centering
\label{tab12}
\vspace{-10pt}
\end{table*}%

\subsection{Baseline for Brute-force Scenario}
To compare with brute-force scenario where an MPI is calculated for each time frame. We have tested LLFF19 \cite{mildenhall2019llff} that includes all views 1-11 in a local fusion manner, and view 12 for testing. It takes 39.5649 seconds to infer MPIs for a single frame, with average PSNR and SSIM 35.41 and 0.95, compared to 31.94 and 0.917 of the Temporal-MPI. Note that the baseline calculates and fuses several static MPIs for each frame, while we only calculate one neural MPI for the entire sequence.

\section{Concluding Remarks}

\subsection{Limitations}

Modeling dynamic scenes is challenging due to complex motions of dynamic objects over time, and specular surface and occlusions on angular domain. Our method makes the first attempt to use a compact temporal representation to reproduce dynamic scenes in time-sequences. Similar to \textit{NeRF20}, our method requires optimization for each scene. Additionally, the output resolution is limited due to limited GPU memory. Furthermore, the rendering quality degrades when the length of sequence increases given default model parameters.  Our approach is also only applicable to dynamic scenes without large camera motions that cause the change of background.

\subsection{Conclusion}
We have proposed a novel dynamic scene representation on top of Multi-plane Image (MPI) with basis learning. Our representation is efficient in computing, thus allowing real-time rendering of dynamics. Extensive studies on public dataset demonstrate the competitive rendering quality and efficiency of our approach. We believe using basis learning for temporal recovery and compression can be applied to the general problem of modeling dynamic contents and not limited to MPI. Using hierarchical encoding method to improve the learning power of MLP on modeling long-time-serial data could be a future extension of our work.

\textbf{Acknowledgments}
The research was supported by the Theme-based Research Scheme, Research Grants Council of Hong Kong (T45-205/21-N).

\bibliographystyle{IEEEbib}
\bibliography{main}

\end{document}